\documentclass[11pt,twoside]{article} 
\usepackage{acta-info}
\usepackage{euler}
\usepackage{float}
\usepackage{algorithm2e}
\usepackage{subcaption}
\usepackage{graphicx}
\newcommand{\vol}{\textbf{}  () }   
\newcommand{\amss}{\amssubj{%
		68T20
}}     
\newcommand{\CRs}{\CRsubj{%
		I.2.6, I.5.1
}}   
\newcommand{\keyw}{\keywords{%
		deep learning, hyperparameter optimization, classification, data mining
}}

\begin{document}
	
	\title{%
		On tuning deep learning models: a data mining perspective
	}
	
	\maketitle
	
	\oneauthor{%
		\href{https://w3.sdu.edu.tr/personel/08926/dr-ogr-uyesi-muhammed-maruf-ozturk/}{Muhammed Maruf \"OZT\"URK} 
	}{%
		\href{http://muhendislik.sdu.edu.tr/bilmuh/en}{Computer Engineering Department, Suleyman Demirel University\\ Isparta, TURKEY}
	}{%
		\href{mailto:muhammedozturk@sdu.edu.tr}{muhammedozturk@sdu.edu.tr}
	}
	

	
	\short{%
		M.M. Ozturk
	}{%
		On tuning deep learning
	}

	\begin{abstract}
		Deep learning algorithms vary depending on the underlying connection mechanism of nodes of them. They have various hyperparameters that are either set via specific algorithms or randomly chosen. Meanwhile, hyperparameters of deep learning algorithms have the potential to help enhance the performance of the machine learning tasks. In this paper, a tuning guideline is provided for researchers who cope with issues originated from hyperparameters of deep learning models. To that end, four types of deep learning algorithms are investigated in terms of tuning and data mining perspective. Further, common search methods of hyperparameters are evaluated on four deep learning algorithms. Normalization helps increase the performance of classification, according to the results of this study. The number of features has not contributed to the decline in the accuracy of deep learning algorithms. Even though high sparsity results in low accuracy, a uniform distribution is much more crucial to reach reliable results in terms of data mining.
	\end{abstract}
	
	
	\section{ Introduction}

	Deep learning (DL) dates back to 1999 \cite{lecun2015deep} when GPU was first developed. It is a sophisticated type of neural network (NN) that has a limited number of hidden layers compared to DL. It has been reviewed in several studies \cite{guo2016deep,voulodimos2018deep,fawaz2019deep} which show to what extent DL has progressed in recent years. DL algorithms have various hyperparameters that are mostly configurable. Some of these hyperparameters are learning rate, hidden layers, and the number of iterations. The number of hyperparameters changes depending on the type of DL. However,    existing approaches generally refer to a specific hyperparameter in order to delve into how the results take shape depending on the research problem. An extensive study explaining the correlation between hyperparameters and performance evaluation in terms of the type of DL algorithms is needed in this field.
	
	DL algorithms are developed based on shallow neural networks that have a limited number of hidden layers. In this respect, the low computational capacity of shallow neural networks limits their functional potential. To solve that problem, a great number of hidden layers~ is designed to give DL models impressive computational capacity \cite{schmidhuber2015deep} compared to the traditional neural networks. DL algorithms are mainly executed for specific tasks, including classification \cite{kussul2017deep,chen2014deep,chan2015pcanet} and regression \cite{held2016learning,wang2017deep,suk2017deep}. The purpose of those applications is to reduce errors or to increase the accuracy of classification experiments. On the other hand, regression aims at reducing mean squared error on the basis of a regression rule.
	
	Hyperparameter optimization and tuning parameters are used interchangeably in related studies \cite{li2017hyperband,scornet2017tuning,wang2016novel}. Attaining an optimal hyperparameter set means that the most suitable configuration yielding the best performance has been found. However, optimal hyperparameters change depending on the structural properties of the data set. For example, a highly balanced data set could lead to overfitting. More importantly, dividing a data set into training, validation, and testing parts makes the optimization results method more reliable.

	To date, hyperparameter optimization strategies implemented to DL algorithms have been successful. Notwithstanding the success of these strategies, a comprehensive study that could make a major advance in understanding to what extent hyperparameter optimization is depended on the type of DL is needed in this field. To address this issue, this paper~reveals~ which ways are the best to conduct hyperparameter tuning for DL methods. To that end, four DL algorithms including deep belief network (DBN), recurrent neural network (RNN), feed-forward neural network (FFNN), and stacked autoencoder (SAE) are analyzed. Moreover, some data mining strategies are involved in the experiment to enhance the comprehensiveness of the study. 
	
	The remainder of the paper is organized as follows: Section 2 describes DL models and summarizes tuning studies using DL. A data processing perspective with regard to DL is presented in Section 3. The most common hyperparameter search methods are given in Section 4. Experimental settings are detailed in Section 5. The results of the experiment are elaborated in Section 6. Last, overall conclusions are drawn and discussed in Section 7.
	\section{ Deep learning}
	\subsection{Deep belief network}
	A DBN is constructed via various stacked RBMs or Autoencoders \cite{hinton2006fast}. All the layers of DBN are directed except for the top two layers. An undirected connection provides an associative memory. Further, unsupervised learning can be conducted through a DBN as well as classification. Observed variables are extracted from the lower layers.~ Visible units in a DBN take input as binary or real data. Figure 1 shows a general structure of DBN. $h_{0}$ refers to the lowest layer which takes input data. The elements in that layer are called visible units~\emph{VU.~}Hidden layers~\emph{h~}and~ \emph{VU~}establish the    model according to Equation \ref{equation1}. Here conditional distributions of~ \emph{VU~}are\emph{~}represented with $M(h^{i}|h^{i+1})$. For top level of DBN, joint distribution is denoted with~ \textit{M(h\textsuperscript{n-1},h\textsuperscript{n})}.

	\begin{figure}[H]
		\begin{center}
			\includegraphics[width=0.4\columnwidth]{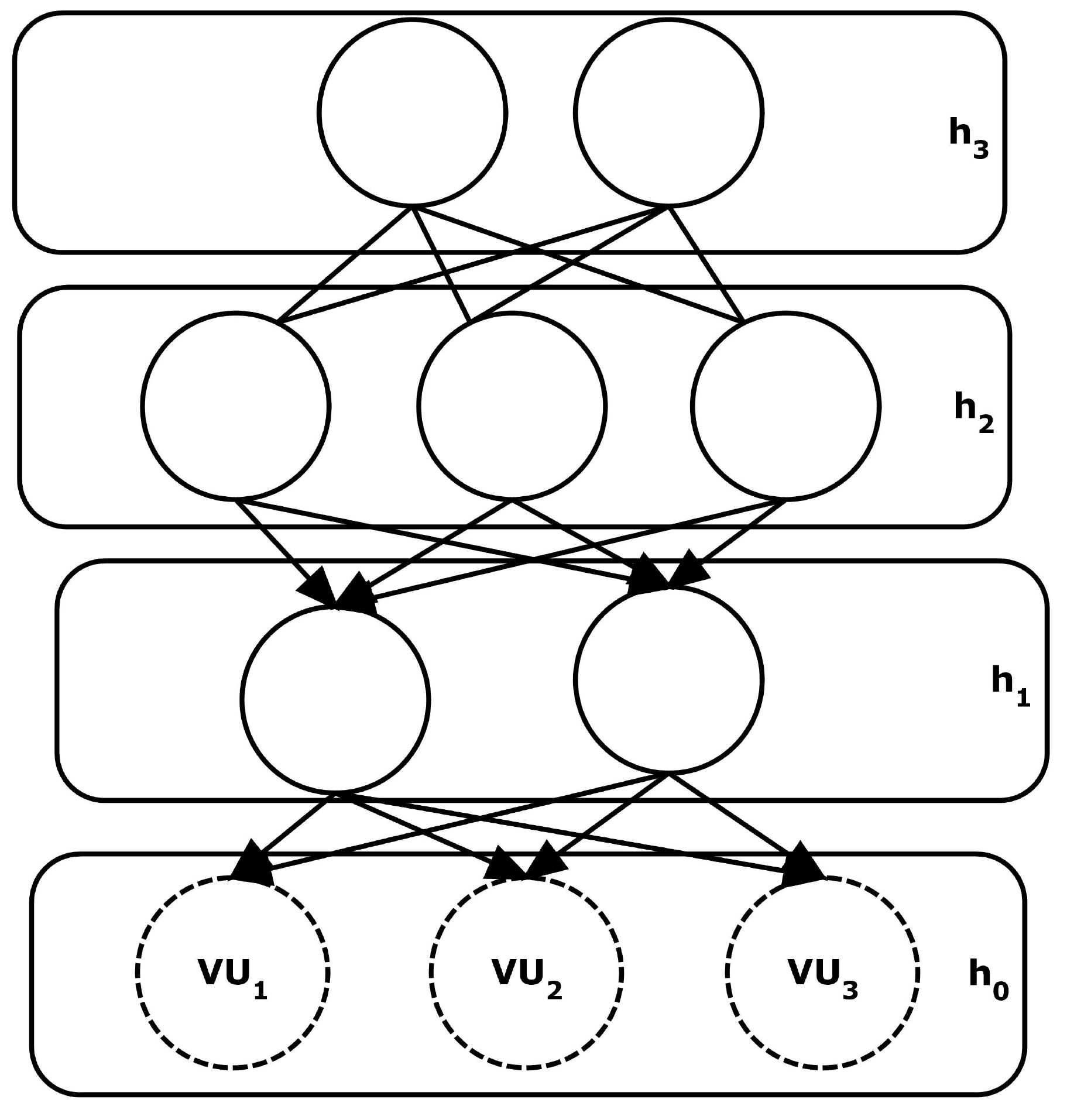}
			\caption{{An overview of DBN.
					{\label{646178}}%
			}}
		\end{center}
	\end{figure}
	\begin{equation}
	M(VU,h^{1},...,h^{n})=\prod_{i=0}^n M(h^{i}|h^{i+1}).M(h^{n-1},h^{n})
	\label{equation1}
	\end{equation}
	
	\subsection{Feed-forward neural network}
	One of the fundamental types of DL algorithms is FFNN which has a fully-connected structure \cite{tomar2014towards}. Unlike the convolutional neural network (CNN), FFNN does not have a convolutional layer. On the other hand, CNN has a backward propagation in the convolution layer. That property makes CNN a good alternative in image classification in which a great number of images are filtered via convolutional layers. 
	
	Fully-connected structure of FFNN creates a significant computational burden for machine learning tasks. Performing a pruning on the DL network may alleviate that burden. The hidden number of layers and hidden units in each layer are of great importance to solve a given problem. Although a wrong configuration of hyperparameters sometimes gives promising results, it may lead to overfitting. Allocating a good representative validation set is a possible solution to avoid overfitting. Getting more training data is another option that requires a large memory.

	\subsection{Recurrent neural network}
	
	RNN provides a powerful alternative for predicting the patterns of sequential data \cite{graves2013speech}. Text sequence prediction and speech recognition are the main application areas of RNN. It gives each output as input in the hidden layers which generates a memorizing mechanism. RNN is much more useful in time series prediction because each hidden layer remembers its previous input. If a classification is performed with RNN, it assumes that features are somehow correlated. For that reason, training RNN takes far more time than that of other types of DL. If previous state and input state are represented with $p_{s}$ and $i_{s}$, respectively. Current state $c_{s}$ of an RNN can be formulated with the following function:
	\begin{equation}
	c_{s}=f(p_{s},i_{s})
	\label{cs}
	\end{equation}
	where $c_{s}-1=p_{s}$. If tanh is chosen to establish activation function, the formula given below describes the activation formula:
	\begin{equation}
	c_{s}=tanh(w_{nn}.p{s}+w_{in}.i_{s})
	\label{activation}
	\end{equation}
	where $w_{nn}$ is the weight of recurrent neuron and $w_{in}$ is the wight of input neuron. Equation \ref{y} describes the general output of a RNN in which $w_{o}$ denotes the weight of output layer.
	\begin{equation}
	y=w_{o}.c_{s}
	\label{y}
	\end{equation}

	\subsection{Stacked autoencoder}

	An SAE consists of a lot of autoencoders that each of them has a single layer \cite{qi2014robust}. An autoencoder has two parts: encoder and decoder. A high dimensional-input is coded in an encoder. On the other hand, a decoder transforms a coded input into a high-dimensional input.  More specifically, SAE offers a mechanism for stacking autoencoders, thereby enabling compression for input data. The number of inputs of autoencoders decreases as the level of SAE increases. An SAE thus is very effective for applications in which data compression is needed. 
	
	\subsection{Hyperparameter tuning of deep learning}
	
	\begin{table}
		\setlength{\tabcolsep}{3pt}
		\centering
		\caption{Overview of some hyperparameters of DL. } 
		\scalebox{0.6}{
			\begin{tabular}{l rrrr} 
				\hline 
				
				\textbf{Name} & \textbf{Description} & \textbf{Method} \\ [1.5ex]
				\hline 
				iteration & the number of iterations over training data to train the model & DBN, FFNN, RNN, SAE \\ 
				batch size & the batch size used for training & DBN, FFNN, RNN, SAE \\ 
				hidden dropout & drop out fraction for hidden layer & DBN, FFNN, SAE \\ 
				visible dropout & drop out fraction for input layer & DBN, SAE \\ 
				learning rate & learning rate for gradient descent & DBN, FFNN, RNN, SAE \\ 
				hidden dim & dimensions of hidden layers or number of units of hidden layers & DBN, FFNN, RNN, SAE \\ 
				\hline 
			\end{tabular}
		}
		\label{hyperparameters}
		
	\end{table}
	This section details DL studies including hyperparameter optimization. The brief gathered from the most recent studies presents the current status of hyperparameter optimization in DL.
	
	Some parameters are considered when optimizing a DL algorithm as follows: learning rate, loss function, mini-batch size, the number of training iterations, and momentum. 
	Table \ref{hyperparameters} presents hyperparameters which are common in DL tuning studies. The column namely "Method" includes DL algorithms that are involved in the experiment. Despite the fact that the number of hyperparameters of a DL algorithm can be up to 11, some experimental constraints such as time and memory confine the bound of a tuning experiment.
	
	Ilievski et al. \cite{ilievski2017efficient} developed a new algorithm namely HORD for optimizing DL models. The evaluation performed with 200 iterations showed that HORD outperformed the other three comparison methods in terms of the validation error. HORD is much more suitable for high-dimensional problems and it runs nearly six times faster than its counterparts. 
	
	Yoo \cite{yoo2019hyperparameter} asserts that nonlinear search methods find optimal hyperparameters faster and with relatively less complexity compared to random Search. He also detected that the success of derivative-free methods can be improved through nonlinear search methods. If a parallelization is required in hyperparameter optimization, some methods such as Bayesian optimization can not meet expectations of optimization above a certain level. To address that problem, Loshchilov and Frank Hutter \cite{loshchilov2016cma} proposed a novel method called CMA-ES that does not include any derivative operation. The method significantly alleviates the computational burden thanks to its design which is fully compatible with parallelization. To increase the speed of DL networks, Domhan et al. \cite{domhan2015speeding} developed a technique based on controlling the learning curve. The success achieved by that method is two times greater than that of the state-of-the-art methods. In \cite{diaz2017effective}, a derivative-free method was presented for hyperparameter optimization. Despite the fact that it does not always reach global optimum, three benchmarks yielded high accuracy for the experimental data set. Yaseen et al. \cite{yaseen2018deep} stressed that setting the learning rate of a DL model to low-value results in successful classification for video data set. In addition to engineering applications, hyperparameter optimization was also evaluated in theoretical studies. For instance, Baldi and Sadowski \cite{baldi2015enhanced} utilized a Bayes optimization algorithm to detect the decay in the Higgs Boson particle. In the detection of decay, an improvement of up to 25\% was able to achieve for both shallow and deep networks. Traditional methods such as Bayesian optimization require a piece of expert knowledge. To solve this problem, Nelder-Mead was tried \cite{ozaki2017effective}. According to the obtained results, Nelder-Mead convergences faster than other methods. Young et al.'s work \cite{young2015optimizing} proposes that using historical results along with an evolutionary approach produces more reliable results than those of random search. However, the method developed by them needs to be validated as it was only tested on one type of DL network.  
	\subsubsection{ Learning rate}
	Tuning the learning rate of a DL algorithm requires an automatic control mechanism to alleviate the computational burden. To that end, Duchi et al. \cite{duchi2011adaptive} devised an adaptive online learning algorithm called ADAGRAD which establishes an inverse relationship between the occurrence of features and learning rate. However, ADAGRAD is very sensitive to the initialization of parameters of gradients that leads to giving low learning rate for some parts of training. ADADELTA \cite{zeiler2012adadelta} addressed that issue, thereby controlling first-order information. Although ADADELTA outperformed stochastic gradient descent (SGD) and momentum in test error, it needs to be re-designed to trivial computations. Zhao et al. \cite{zhao2019research} utilized an energy neuron model to decide the learning rate of DL by analyzing features. Further, they pointed out that there are tradeoffs in all couple hyperparameters. To set the learning rate, in \cite{chandra2016deep}, the laplacian score is employed to increase the success of classification. Laplacian score has a great potential to give information about the significance of neurons of DL. In \cite{keskar2015nonmonotone}, cross-validation helped to attain optimistic biases in the learning rate of DL. In addition, a layer-specific adaptive scheme \cite{singh2015layer} was found beneficial to speed up learning in the initial layers of DL. Smith \cite{smith2017cyclical} depicted that using a cyclic learning rate improves the classification accuracy of DL. He also noted that optimal classification result is obtained via a cyclic learning rate in few iterations.
	
	The studies mentioned above mostly recommend comprehensive trials on the learning rate of DL. Further, each adaptive learning scheme for the learning rate is highly dependent on the type of DL model established to conduct a specific machine learning task.
	\subsubsection{Batch size}
	The batch size determines the number of instances that are used in training for each update of model parameters. Employing a large batch size requires a high memory capacity. Therefore, batch size should be optimized in compliance with the machine configuration.
	
	Li et al. \cite{li2018adaptive} developed a batch normalization method called AdaBN for DL models. They concluded that setting a small number of instances for batch size may not yield consistent accuracy. For that reason, a threshold should be set for batch size. Above this value, adding more instances to the batch size does not change the accuracy. Besides reaching a stable accuracy, batch size helps accelerate DL models according to the experiment performed by Liu et al. \cite{liu2018deep}. Santurkar et al. \cite{santurkar2018does} examined the internal covariate shift to observe the advantages of employing batch normalization. They stressed that batch normalization makes gradients of training more predictive that results in faster yet effective optimization. For a robustness analysis of DL, Yao et al. \cite{yao2018hessian} devised a hessian-based experiment on the CIFAR-10 data set. The robustness of a DL model is highly dependent on the batch size according to their experiment. They also depicted that batch size should be kept small. Bjorck et al. \cite{bjorck2018understanding} argue that activations of DL grow at a fast pace if the learning rate is too large. In their study, batch normalization was found as the sole way to prevent the explosion of DL networks. For complex optimization problems, a mini-batch stochastic gradient descent was proposed in \cite{li2014efficient}. The most important facility of the method is that it helps keep the convergence rate at a reasonable level even if the batch size increases significantly. Some researchers preferred to analyze a specific DL model in terms of batch size. For instance, Laurent et al. \cite{laurent2016batch} investigated the effects of optimizing batch size on recurrent neural networks. They proposed to use batch normalization to achieve fast training in recurrent neural networks. Besides fast training, using an optimal batch size reduces the need for parameter updates \cite{smith2017don}.
	\subsubsection{Hidden node and hidden layer}
	Designing many hidden layers in DL models leads to a high memory requirement. Further, such models spend too much time to complete training. To address this problem, Alvarez and Salzmann \cite{alvarez2016learning} proposed an approach to determine the number of neurons in layers of a DL network. The method achieved a great speedup at testing. Some works are designed for determining both the number of neurons and the number of hidden layers. Thomas et al. \cite{thomas2016discovery} performed such an experiment for feed-forward neural networks. They were able to achieve high accuracy in classification. Another study was performed by Xu et al. \cite{xu2016learning}. They revealed that despite the fact that using a great number of hidden layers sharpens the learning model with respect to the training accuracy, they remarkably increase the effort needed for training. Morales \cite{kuri2017closed} investigated multi-layer perceptron for finding the optimal number of hidden neurons. They concluded from a comprehensive experiment that determining the correct number of neurons is highly correlated with the size of the training data.
	neurons. They concluded from a comprehensive experiment that determining the correct number of neurons is highly correlated with the size of the training data.
	\subsubsection{Dropout}
	Dropout is a configuration parameter that is used in input and output layers as a rate for ignoring neurons. Dropout helps avoid overfitting to generalize a DL model.
	
	Some researchers argue that dropout should be set according to the underlying mechanism of DL being used. Ba and Frey \cite{ba2013adaptive} proposed an adaptive algorithm for dropout. Their method achieved a remarkable reduction in classification error when using shallow networks. Classification performance of using adaptive dropout was also investigated by Kingma et al. \cite{kingma2015variational}. They depicted that choosing an adaptive dropout helps reduce classification error, remarkably. Bayesian approaches were mostly utilized in studies dealing with dropout rates. For instance, in \cite{zhuo2015adaptive}, a Bayesian dropout learning method was proposed classification. The experiment showed that employing an adaptive dropout rate reduces the effort allocated to perform tuning. Phum et al. \cite{pham2014dropout} analyzed the effects of tuning dropout of recurrent neural networks. Performing a tuning on dropout not only reduces classification error but also provides a relatively easy way to perform tuning. Ko et al. \cite{ko2017controlled} pointed out that dropout should be set according to its corresponding network model. Using a dropout range between 0.4 and 0.8 was strongly advised by them to keep test error low. Zhang et al. \cite{zhang2018adaptive} utilized a distribution function to configure dropout. They found adaptive dropout learning to have a high potential to conduct big data learning in IoT.
	\section{Data processing}
	This section is divided into two subsections detailing DL studies in terms of data cleansing and data normalization. Since data preparation is of great importance for shallow neural networks \cite{lopez2019shallow}, as well as DL \cite{pal2016preprocessing}, two essential data processing methods are summarized to explain their relationships with DL. 
	
	\subsection{Data cleansing}

	Chuck et al. \cite{chuck2017statistical} proposed a data cleansing algorithm based on low-confidence correction for the planar part extraction task. The method was able to achieve up to 10\% decrease in training loss. They also argue that inconsistency in the results obtained via machine learning algorithms is not originated from wrong formatting, but rather due to misinterpretation of data. Noisy data distribution can be detected via modified deep learning models. Sukhbaatar and Fergus \cite{sukhbaatar2014learning} proposed a deep learning model for noisy data. Their method was trained on clean data to predict noise distribution. Noisy data was found beneficial to reduce the error rate of training in that study. In \cite{vo2017harnessing}, a kernel mean matching technique was devised to learn from noisy data. It achieved a good generalization along with an improved classification using FlickR images. Zhang and Sabuncu \cite{zhang2018generalized} argued that mean absolute error has some drawbacks to evaluate the performance of a deep neural network. When using complicated data sets, the mean absolute error creates a remarkable difficulty in training as it provides robustness against noisy labels. To address this problem, they proposed two loss functions for classification. They yielded high accuracy on the instances featuring noisy labels. Bekker and Goldberger \cite{bekker2016training} proposed a new deep learning algorithm which does not need any clean data to perform training on noisy data. Their method showed great resistance against high noise fraction. Massouh et al. \cite{massouh2017learning} described external noise as a wrong label which is not available in the instances. They depicted that CNN shows higher robustness to external noise than to internal noise. Choi et al. \cite{choi2018effects} employed CNN to tagging music. They argue that the tag-wise performance of a data set shows the noisiness of it. Wu et al. \cite{wu2018light}  designed a light CNN for face recognition. It works faster than traditional CNN's due to its feature map operation that makes CNN relatively small. They also proposed a bootstrapping to cope with noisy labels in images. Li et al. \cite{li2019object} devised a CNN which yields high accuracy on data sets where noise distribution of them is ambiguous. 
	Most of the researches concerning deep learning with noisy data is focused on label noise. Instead, noisy data problems should be addressed regarding other data problems such as sparsity. Further, there is a need to develop learning techniques in the context of the experiments in which both the labels and the features have noisy points.
	
	\subsection{Data normalization}

	In deep learning methods, data normalization can be divided into three categories: batch normalization, input weight normalization, and raw input normalization. Liu et al. \cite{liu2018deep} proposed a batch normalization technique for character recognition. In their experimental setup, a batch normalized layer was added to a fully connected deep neural network to improve the generalization of their method. They detected that employing ReLU brings the results 12 times faster than sigmoid. In \cite{wang2017effectiveness}, data augmentation was utilized to increase the accuracy of classification. To that end, an augmentation network was established in training. According to the results of the experiment, data augmentation helps increase the success of classification in which there is a lack of training data.  
	
	Data augmentation was also applied to predicting gait sequences \cite{wang2017effectiveness}. Performing training and testing on different data sets by using either real data or synthetic data results in low accuracy. To address that problem, training the model using mixed data is a good way to achieve 95\% of accuracy. Zhong et al. proposed a data augmentation method for various recognition tasks of image recognition. Their method randomly selects a rectangle region in an image and then changes the values of pixels of that rectangle by using random values. Even though their method increased the accuracy of CNN up to 3\%, some questions remained unanswered. For example, does expanding an image with arbitrary pixels rather than changing the values of the pixels of a specific region in it give promising results? Moreover, the critical question is, to what extent image-quality based value generation is compatible with the CNN designed by \cite{taylor2018improving}. They argue that cropping is the best way to perform data augmentation for CNN. Further, they detected color-jittering as the second successful method for data augmentation. In \cite{bhanja2018impact}, Tanh Estimator was found to be the most effective normalization method for recurrent neural networks. However, to generalize the results, other types of deep learning methods such as stacked autoencoders should be involved in a comprehensive experiment. Passalis et al. \cite{passalis2019deep} proposed a new normalization method called DAIN for deep learning models. They evaluated the method on three types of deep learning algorithms. The method yielded the highest accuracy among the comparison methods as it has an adaptive scheme to perform normalization in which data distribution is analyzed to avoid a fixed scheme. Tran et al. \cite{tran2017bayesian} developed a data augmentation method on the basis of Bayesian reasoning. The method achieved high accuracy on the classification of image data sets. Sound data has the potential to perform deep learning experiments as well as image data. In \cite{salamon2017deep}, data augmentation in sound data sets resulted in a 0.5 increase in mean accuracy. 
	\section{Tuning strategies}

	\subsection{Grid Search}
	
	{\label{389878}}
	Grid search is one of the most common hyperparameter search methods and it searches all the parameter space \cite{hutter2015beyond,ensor1997stochastic}. Unlike random search, it performs exhaustive searching in specific distinct places which depend on the number of types of hyperparameters \cite{bergstra2012random}. Let $T_{1}$ be training set and $T_{2}$ denotes testing set, each configuration of hyperparameter set is trained on $T_{1}$ to test with $T_{2}$. Despite the fact that grid search provides an exhaustive evaluation of hyperparameters, unlike other search methods, it requires a great number of iterations. $d$ denotes the dimension of hyperparameter and $c$ is the possible choice of hyperparameters where $n$ is the number of iterations, the complexity of grid search can be calculated with the following equation:    
	
	\begin{equation}
	O(\dfrac{V}{n} .c^{d}.F(Q,\lambda))
	\label{gridSearch}
	\end{equation}
	
	where $F(Q,\lambda)$ minimizes the criterion which decides when training is suspended. $V$ represents the total number of predictions.
	\subsection{ Derivative-free methods}
	
	This subsection describes four derivative-free optimization methods which are also suitable for hyperparameter optimization. However, we did not include any derivative-free method in the experimental study except Random Search because if the problem is very big, it leads to an exponential increase in the number of function evaluations so that an adaptive parallelization \cite{ozaki2019accelerating} is required.

	\textbf{Random search} prefers to search parameters in various distinct places depending on the number of parameters \cite{probst2019hyperparameters}. If the time allocated for searching hyperparameters is limited, the random search could be a possible solution to perform hyperparameter optimization. Further, parallelization can be easily established in a random search as it does not require communication between workers. The following equation describes the computational complexity of random search:
	
	\begin{equation}
	O(\dfrac{V}{n} .R.F(Q,\lambda))
	\label{randomSearch}
	\end{equation}
	
	where $V$ is the targeted parameter volume for $n$ iterations in $R$ space. 
	
	\textbf{Genetic algorithm.} Three concepts constitute the mechanism of evolution: natural selection \cite{smith2017natural}, mutation \cite{allen1969hugo}, and genetic \cite{dobzhansky1950genetic}. Analyzing these three concepts resulted in genetic algorithm which is useful for any optimization problem. The main objective of the genetic algorithm is to find global optimum by searching a space including many local optima. For that reason, it requires a large number of computations. The genetic algorithm utilizes mutation that leads to varying outcomes in which the problem is not adaptable to differentiable or continuous objective functions.

	\textbf{Bayesian optimization.} Bayes theorem is directly related to Bayesian optimization. It builds a probabilistic model by assuming a past event is a base to evaluate a current event. The approximation of Bayes is called a surrogate function that is used for future sampling. In Bayesian optimization, the acquisition criterion decides which sample will be selected in the next evaluation \cite{rios2013derivative}. 
	
	\textbf{Nelder-Mead optimization}, which is very effective for stochastic responses, was first introduced in 1965 \cite{glaudell1965nelder}. Nelder-Mead performs an iterative computation to complete optimization at low-cost. It aims at minimizing the error $e$ of an objective function $f(x)$. Here $x$ is a member of solution space and $f(x)$ is updated for each response as follows:
	\begin{equation}
	f(x)_{2}=f(x)_{1}+e
	\end{equation}
	
	where $f(x)_{1}$ is the output of the previous iteration that is used to calculate $f(x)_{2}$ new result of next iteration, thereby adding error value $e$.Even though Nelder-Mead converges at a fast pace, it is not effective for high-dimensional problems.
	\subsection*{5 \textbar{} EXPERIMENTAL SETTINGS}
	
	The experiment was performed on a machine having CentOS Linux, 64-bit, Intel(R) Xenon(R) 2.9 GHz, 32 CPU Cores server with 263 GB RAM, and Tesla C1060 graphics processor. The R codes of the study can be downloaded via the link (https://github.com/muhammedozturk/deepLearning).
	
	\subsubsection*{5.1 \textbar{} Data Sets}
	24 data sets were collected from the OpenML platform \cite{OpenML2013} which enables researchers to share their data sets to perform machine learning tasks. All the data sets are for classification experiments. Six data sets have factor values that were converted to numeric to make them suitable for DL algorithms. The crucial point in their conversion is making sure that there is no mathematical model in factor values. Otherwise, the conversion should be conducted by giving relational values to them.   Table \ref{datasets} gives the summary of experimental data sets having a various number of instances which range from 540 to 45312. 
	
	Algorithm 1 was designed to perform preprocessing on the experimental data sets. For $M$ matrix, $n$ is the number of data sets. $FactorAnalysis$ checks the last column of a matrix $M_{i}[,n]$ to determine whether the label column includes factor values. Thereafter, $CountFactor$ calculates the number of factor labels to convert factor values to numeric with the help of $Random$ function. The list called $SList$ is generated to collect sparsity results of the data sets. $lengthC$ represents the number of columns and $lengthR$ is the number of rows. The function called $normalize$  conducts a minmax normalization on each data cell of matrix $M$. Last, the processed data group and sparsity list $SList$ are obtained.
	\begin{algorithm}
		\SetKwData{Left}{left}\SetKwData{This}{this}\SetKwData{Up}{up}
		\SetKwFunction{Union}{Union}\SetKwFunction{FindCompress}{FindCompress}
		\SetKwInOut{Input}{input}\SetKwInOut{Output}{output}
		\Input{Data sets $M_{1},...,M_{n}$}
		\Output{Processed data sets $M_{1},...,M_{n}$}
		\BlankLine
		\For{$i\leftarrow 0$ \KwTo n}{
			$FA$ $\leftarrow$ $FactorAnalysis(M_{i}[,n])$;\\
			\If{$FA$$==$TRUE}{
				$CF$ $\leftarrow$ $CountFactor(M_{i}[,n])$	\\
				$M_{i}[,n]$ $\leftarrow$ $Random(1,CF)$
			}
			$SList$ $\leftarrow$ $sparsity(M_{i})$\\
			$lengthC$ $\leftarrow$ $M_{i}[1,]$ \\
			$lengthR$ $\leftarrow$ $M_{i}[,1]$ \\
			\For{$j\leftarrow 0$ \KwTo $lengthR$}{
				\For{$k\leftarrow 0$ \KwTo $lengthC$}{
					$M_{i}[j,k]$ $\leftarrow$ $normalize(M_{i}[j,k])$ 
				}
			}
		}
		Return ($M_{1},...,M_{n}, Slist$)
		\caption{Preprocessing for deep learning.}\label{DFFNNA}
	\end{algorithm}
	\begin{table}
		\setlength{\tabcolsep}{3pt}
		\centering
		\caption{Summary of the experimental data sets. NF=$>$number of features, NI=$>$number of instances, RF=$>$T (true), F (false):indicates whether the data set require an operation for converting factor values to numeric.} 
		\scalebox{0.8}{
			\begin{tabular}{l rrrrr} 
				\hline 
				
				\textbf{Name} & \textbf{Description} & \textbf{NF} & \textbf{NI} & \textbf{RF}  \\ [1.5ex]
				\hline 
				bank-marketing\# & includes marketing campaigns of a banking institution & 16 & 45211 & T\\ 
				blood-transfusion\# & this data is of explicit use in classification & 4 & 748 & F\\ 
				climate-simulation\# & includes failure analysis of simulation crashes in climate models & 20 & 540 & F\\ 
				credit-g\# & includes credit risks of people for classification & 20 & 1000 & T\\ 
				diabetes-37\# & includes some diabet test results & 8 & 768 & F\\ 
				tic-tac-toe\# & includes encoding information of tic-tac-toe game & 9 & 958 & T\\ 
				electricity\# & includes summary information about electricity consumption & 8 & 45312 & F\\ 
				gina-agnostic \# & includes values of handwritten digit recognition & 970 & 3469 & F\\ 
				hill-valley\# & every instance in this data set represent a two-dimensional graph & 100 & 1212 & F\\ 
				ilpd\# & includes some patient records & 10 & 583 & T\\ 
				kr-vs-kp\# & includes chess game records & 36 & 3196 & T\\ 
				madelon\# & an artificial data set for classification & 500 & 2600 & F\\ 
				monks-problems-1\# & contains some values of an algorithm selection problem & 6 & 556 & F\\ 
				monks-problems-2\# & second version of monks-problems & 6 & 601 & F\\ 
				monks-problems-3\# & third version of monks-problems & 6 & 554 & F\\ 
				mozilla4\# & includes information about a conditional model & 5 & 15545 & F\\ 
				musk\# & includes values of a molecule classification & 162 & 6598 & T\\ 
				nomao\# & this data set can be used for location classification & 118 & 34464 & F\\ 
				ozone-level-8hr\# & includes an analysis for predicting ozone level & 72 & 2534 & F\\ 
				phoneme\# & includes information about acoustic observation & 5 & 5404 & F\\ 
				qsar-biodeg\# & this data set is prepared for chemical classification & 41 & 1055 & F\\ 
				scene\# & image recognition data set & 296 & 2407 & F\\ 
				steel-plates-fault\# & includes fault types of plates & 33 & 1941 & F\\ 
				wdbc\# & includes characteristics of the cell nuclei of breast images & 30 & 569 & F\\ 
				\hline 
			\end{tabular}
		}
		\label{datasets}
	\end{table}
	\subsubsection*{5.2 \textbar{} Configurations for training and testing}
	A two-sided classification experiment is designed to evaluate the results: one side is a learning rate based comparison of DL algorithms, the other comprises accuracy evaluations of DL algorithms in terms of some data mining tasks. 
	
	The value space ranging from 0.005 to 0.823 is set for the learning rate. The step size is 0.05 that resulted in 208 values by adding some new values to space by ignoring that step size. Each data set is divided into 70\% for training and 30\% for testing. Since we observe the effect of a specific hyperparameter on the accuracy, a validation set is not considered for the first side of the experiment. Mean accuracy results are then obtained via 10*10 cross-validation.
	
	The second side of the experiment is to evaluate the accuracy of DL algorithms in terms of data mining tasks. To that end, sparsity and normalization analyzes are involved in the experiment. For a matrix $M$, sparsity is calculated via the following formula:
	\begin{equation}
	S=\dfrac{M_{z}}{M_{t}}
	\label{sparsity}
	\end{equation}
	
	where $M_{z}$ is the number of zero values, $M_{t}$ is the number of total elements, and $S$ refers to the sparsity rate. Minmax normalization, which is applied on each feature of the data set, is calculated using (\ref{normalization}) as follows:
	\begin{equation}
	M[i,j]=\dfrac{M_{[i,j]}-min(M_{[,j]})}{max(M_{[,j]})-min(M_{[,j]})}
	\label{normalization}
	\end{equation}
	which yields a normalized $M$. Here $j$ denotes the feature and $i$ refers to the instance.
	
	The data sets are divided into three parts including training, validation, and testing sets for the second side of the experiment. 70\% of each data set is used for training, 15\% of remaining parts are employed for validation that shows the degree to which configuration of hyperparameter sets fits training. The last 15\% of remaining parts are used for testing which assesses the general performance of the classifier.
	
	We ran random search and grid search algorithms for three hyperparameters as follows: learning rate (0.1,0.2,...,0.9), batch size (10,20,...,100), and number of hidden nodes (1,2,...,10) are for FFNN; learning rate (0.1,0.2,...,0.9), batch size (10,20,...,100), and number of hidden nodes (1,2,...,10) are for DBN; learning rate (0.1,0.2,...,0.9), numepochs (10,20,...,100), and dimension of hidden layers (1,2,...,10) are for RNN; learning rate (0.1,0.2,...,0.9), batch size (10,20,...,100), and number of hidden nodes (1,2,...,10) are for SAE. Drop out rates for visible and hidden layers are set to 0 as default.
	\section{Results}
\begin{figure}[H]
		\begin{subfigure}{.5\textwidth}
			\centering
			\scalebox{1.3}{
				\includegraphics[width=.8\linewidth]{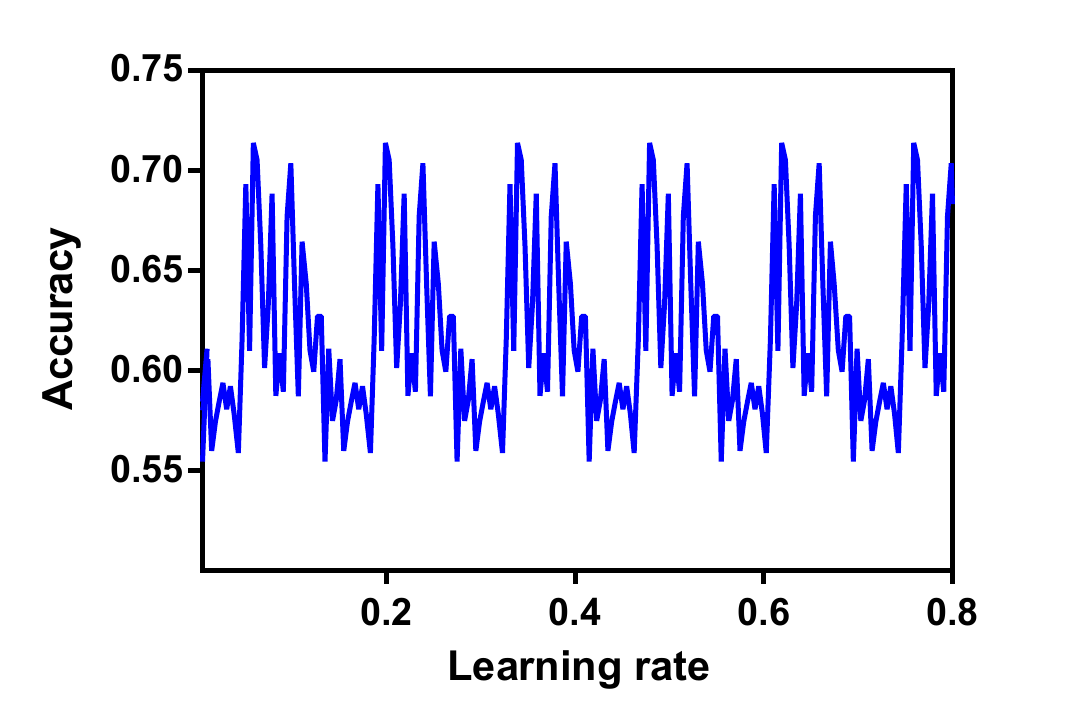}}
			\caption{FFNN}
			\label{ffnn}
		\end{subfigure}
		\begin{subfigure}{.5\textwidth}
			\centering
			\scalebox{1.3}{
				\includegraphics[width=.8\linewidth]{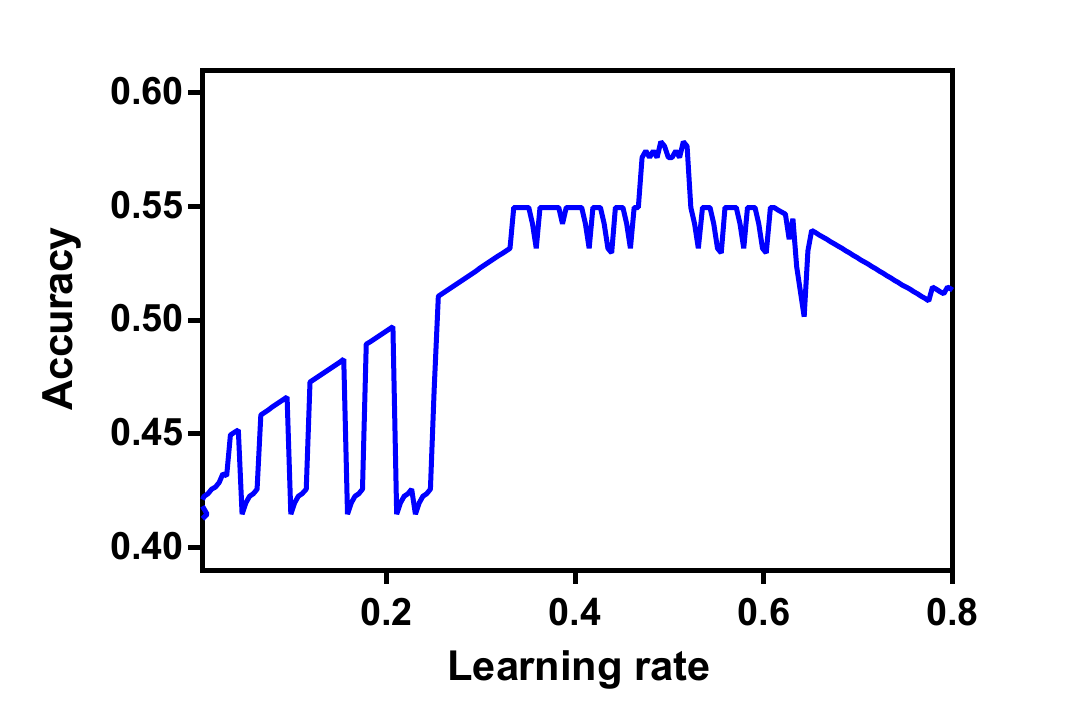}}
			\caption{RNN}
			\label{rnn}
		\end{subfigure}
		\begin{subfigure}{.5\textwidth}
			\centering
			\scalebox{1.3}{
				\includegraphics[width=.8\linewidth]{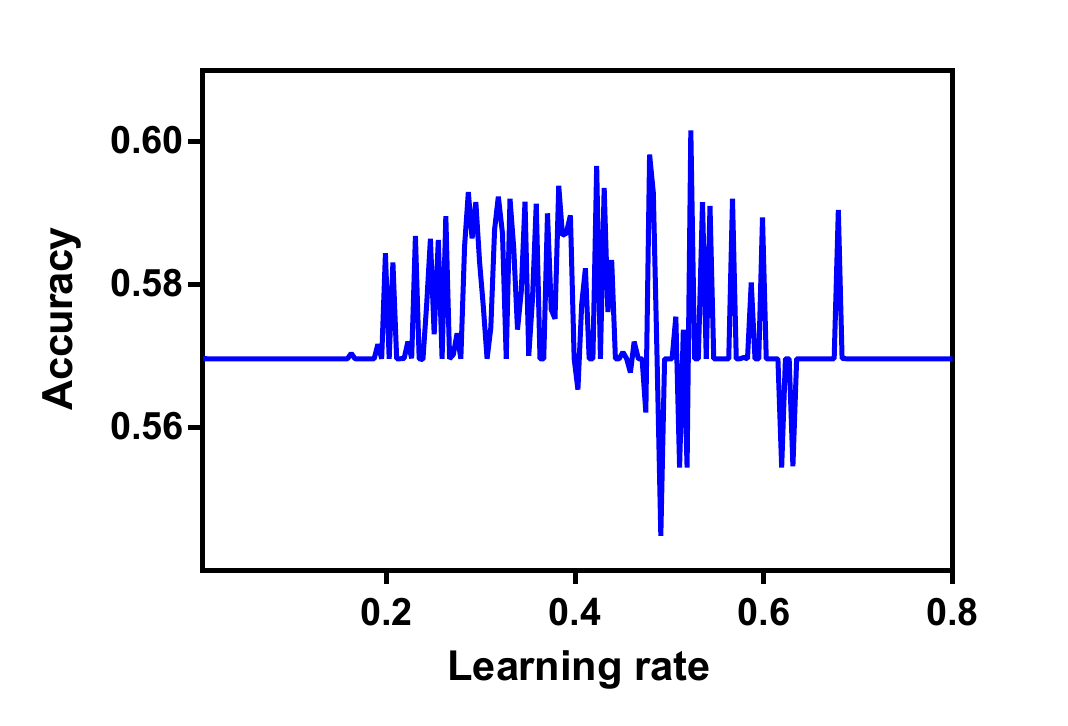}}
			\caption{SAE}
			\label{stacked}
		\end{subfigure}
		\begin{subfigure}{.5\textwidth}
			\centering
			\scalebox{1.3}{
				\includegraphics[width=.8\linewidth]{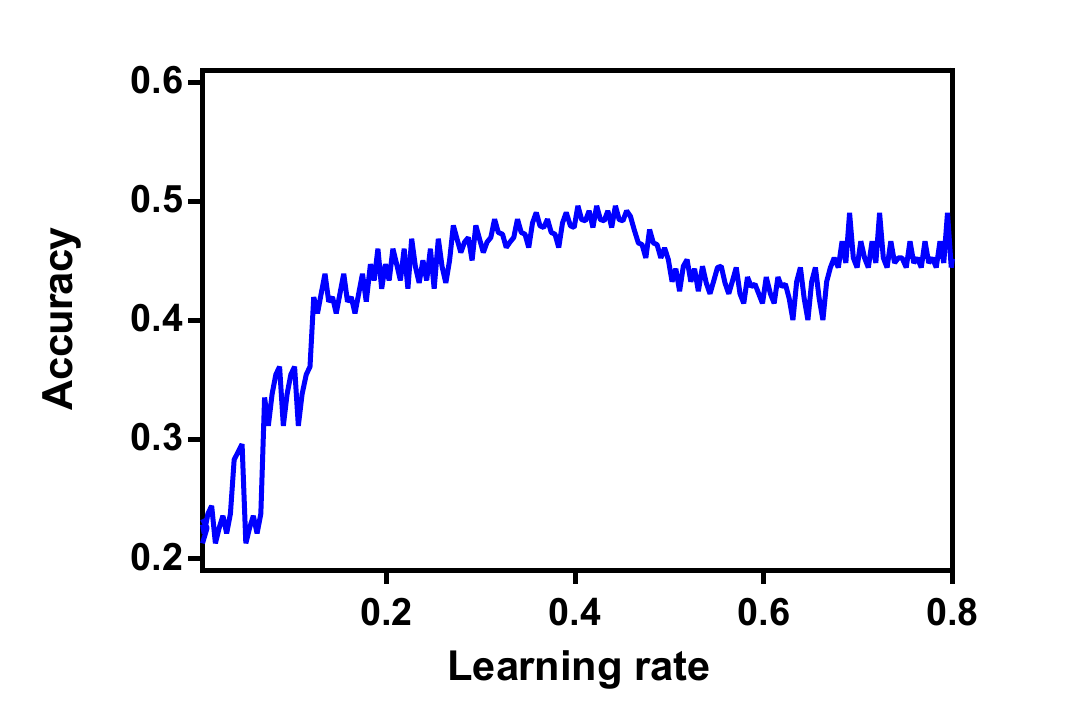}}
			\caption{DBN}
			\label{dbn}
		\end{subfigure}
		\caption{Accuracy changes of DL algorithms depending on the learning rate. The results were validated with Kruskal Wallis Test which performs a non-parametric analysis between the groups. That test does not perform matching to pursue statistical analysis. The results are significantly different (p$<$0.05) according to the Kruskal Wallis Test (reject the populations have the same distributions $H_{0}$). }
		\label{lrGraph}
	\end{figure}

	The change of accuracy values depending on the learning rate is given in Figure \ref{lrGraph}. FFNN has a cyclic and highest learning rate among the comparison methods. Finding a reasonable learning rate is relatively easy in FFNN due to its form of repetitive accuracy. RNN starts converging at 0.47 of the learning rate so that setting maximum learning rate to 0.51 is reasonable for that algorithm. It is clearly seen from Figure \ref{stacked} that training peaks at 0.52 of the learning rate for SAE. However, managing the ups and downs of learning rate of SAE is easy compared to the other methods. Even though DBN yields a stagnant performance in changing learning rates, it has the lowest accuracy among those algorithms. In addition to this, DBN is not sensitive to learning rates which are greater than 0.2. In that case, DBN becomes flexible in which a wide range of learning rate is employed to train a neural network.
	
	DBN requires less time and effort compared with FFNN. Further, DBN achieves higher value of accuracy than FFNN has. A fully connected feed forward neural network needs to be pruned in pre-training. In that way, FFNN can be made fast enough. Moreover, a class-imbalance analysis helps decide which part of data set is suitable for training. An oversampling or undersampling operation may be performed to address that issue. 
	
	\begin{table}
		\setlength{\tabcolsep}{3pt}
		\centering
		\caption{Sparsity rates of experimental data sets.} 
		\scalebox{0.4}{
			\begin{tabular}{l rr} 
				\hline 
				
				\textbf{Name} & \textbf{S}   \\ [1.5ex]
				\hline 
				bank-marketing\# & 0.3101071\\ 
				blood-transfusion\# & 0.001336898\\ 
				climate-simulation\# & 0\\ 
				credit-g\# & 0.07585714\\ 
				diabetes-37\# & 0.1103877\\ 
				tic-tac-toe\# & 0.2066806\\ 
				electricity\# & 0.06653621\\ 
				gina-agnostic \# & 0.689833\\ 
				hill-valley\# & 0.004950495\\ 
				ilpd\# & 0\\ 
				kr-vs-kp\# & 0.00189426\\ 
				madelon\# & 7.676954e-07\\ 
				monks-problems-1\# & 0.07142857\\ 
				monks-problems-2\# & 0.09389113\\ 
				monks-problems-3\# & 0.06859206\\ 
				mozilla4\# & 0.1706229\\ 
				musk\# & 0.00775547\\ 
				nomao\# & 0.01575731\\ 
				ozone-level-8hr\# & 0.01229849\\ 
				phoneme\# & 0\\ 
				qsar-biodeg\# & 0.4520876\\ 
				scene\# & 0.02767761\\ 
				steel-plates-fault\# & 0.2011243\\ 
				wdbc\# & 0.004422019\\ 
				\hline 
			\end{tabular}
		}
		\label{sparsityRates}
	\end{table}

	The validation part is generally utilized to attain a reasonable configuration of hyperparameters. A generalization is thus achieved in the testing phase. However, using a precise ratio to decide the size of training-validation-testing parts may lead to unimproved performance. To solve that problem, in the iterations of cross-validation, division rates of those parts might be changed to achieve a reasonable generalization in the validation phase.

	RNN is generally employed for sequential data models such as text and speech prediction. RNN requires much time when there are a lot of features as it is sensitive to the number of features. In the experiment, the data set namely gina-agnostic has 970 features. The time passed to complete the grid search of RNN for gina-agnostic is 10 times greater than those of other data sets.
	
\begin{figure}
		\centering\scalebox{0.6}{
			\includegraphics{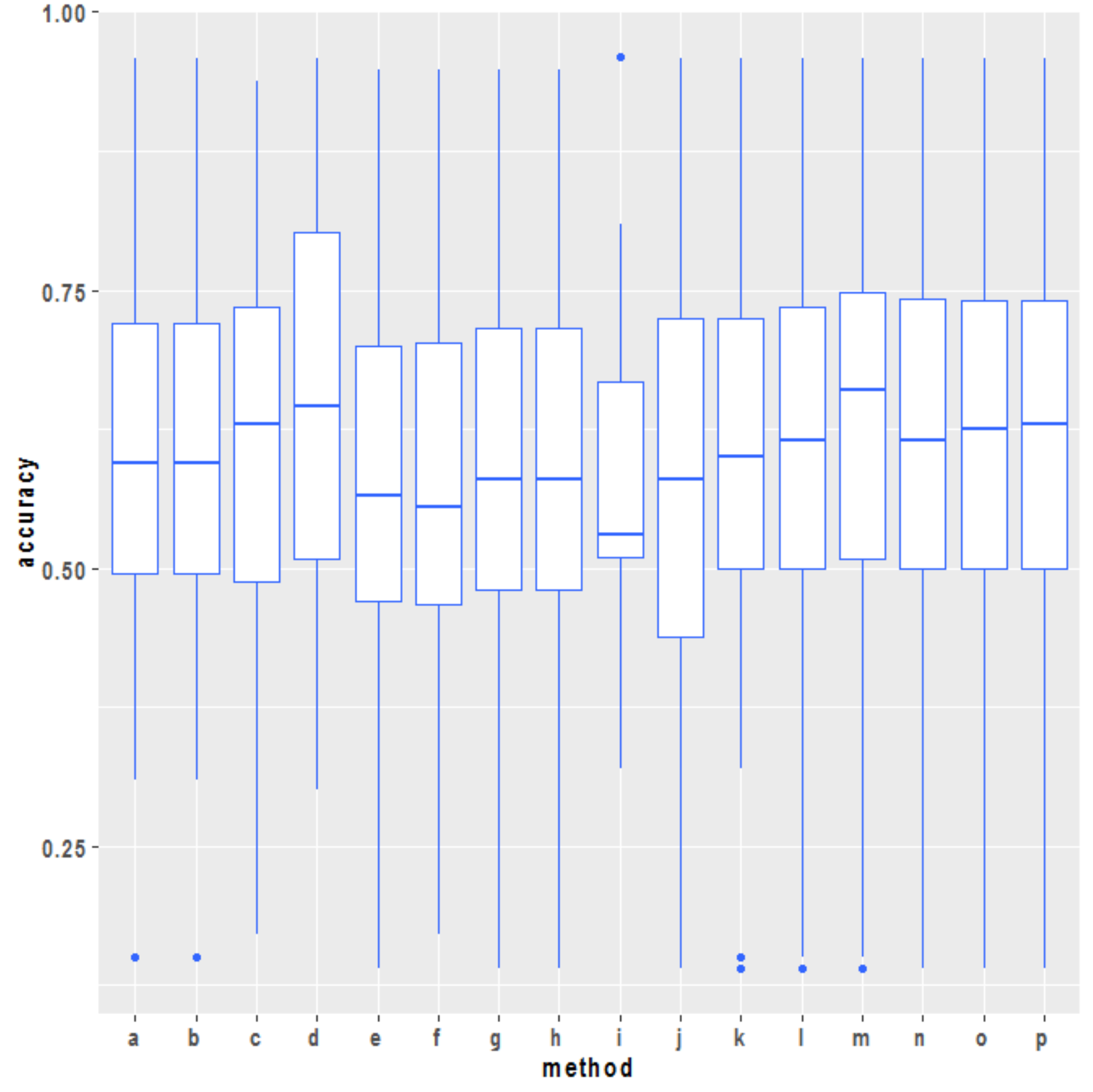}}
		\caption{Box-plots illustrating the effects of normalization on hyperparameter tuning. a:DBN-grid search with original data set, b:DBN-grid search with normalized data set, c:DBN-random search with original data set, d:DBN-random search with normalized data set, e:FFNN-grid search with original data set, f:FFNN-grid search with normalized data set, g:FFNN-random search with original data set, h:FFNN-random search with normalized data set, i:RNN-grid search with original data set, j:RNN-grid search with normalized data set, k:RNN-random search with original data set, l:RNN-random search with normalized data set, m:SAE-grid search with original data set, n:SAE-grid search with normalized data set, o:SAE-random search with original data set, p:SAE-random search with normalized data set,}
		\label{boxauc}
	\end{figure}

	The results of two types of hyperparameter search methods are presented in Figure \ref{boxauc} for four DL algorithms. In total, 16 box plots are obtained by adding the results obtained via normalized data sets. The accuracy of the random search is 5.34 higher than that of the grid search. It can be seen from Figure \ref{boxauc} that normalization does not adversely affect the success of DL algorithms. On the contrary, normalization has a favorable effect on the success of DL in general. Normalization has created the highest increase in the accuracy of RNN. From this point of view, we can conclude that RNN is the most compatible with normalization. Meanwhile, it is worth noting that RNN needs all the features of the testing set to predict sequential data in which an effort-intensive operation is performed. Normalization has decreased the success of stacked autoencoders. It is rather concerned with the dimension of training data instead of the scale. The decline in the success of grid search in SAE may have originated from the direct use of feature space. Normalization has only created an adverse effect on the grid search of SAE given in Figure \ref{boxauc}-(m,n). An SAE is inherently able to reduce the number of inputs as the level of autoencoders increase. It is trivial to perform normalization when using an SAE in which the deviation is not remarkably high. As such, it is extremely important to avoid normalization in SAEs to sustain a successful classification. For all the data sets, random search gives relatively better results than the grid search. However, at that point, the search methods used in this study should be validated with different machine learning tasks such as regression.

\begin{figure}[H]
		\begin{subfigure}{0.5\textwidth}
			\centering
			\scalebox{0.5}{
				\includegraphics[height=5in]{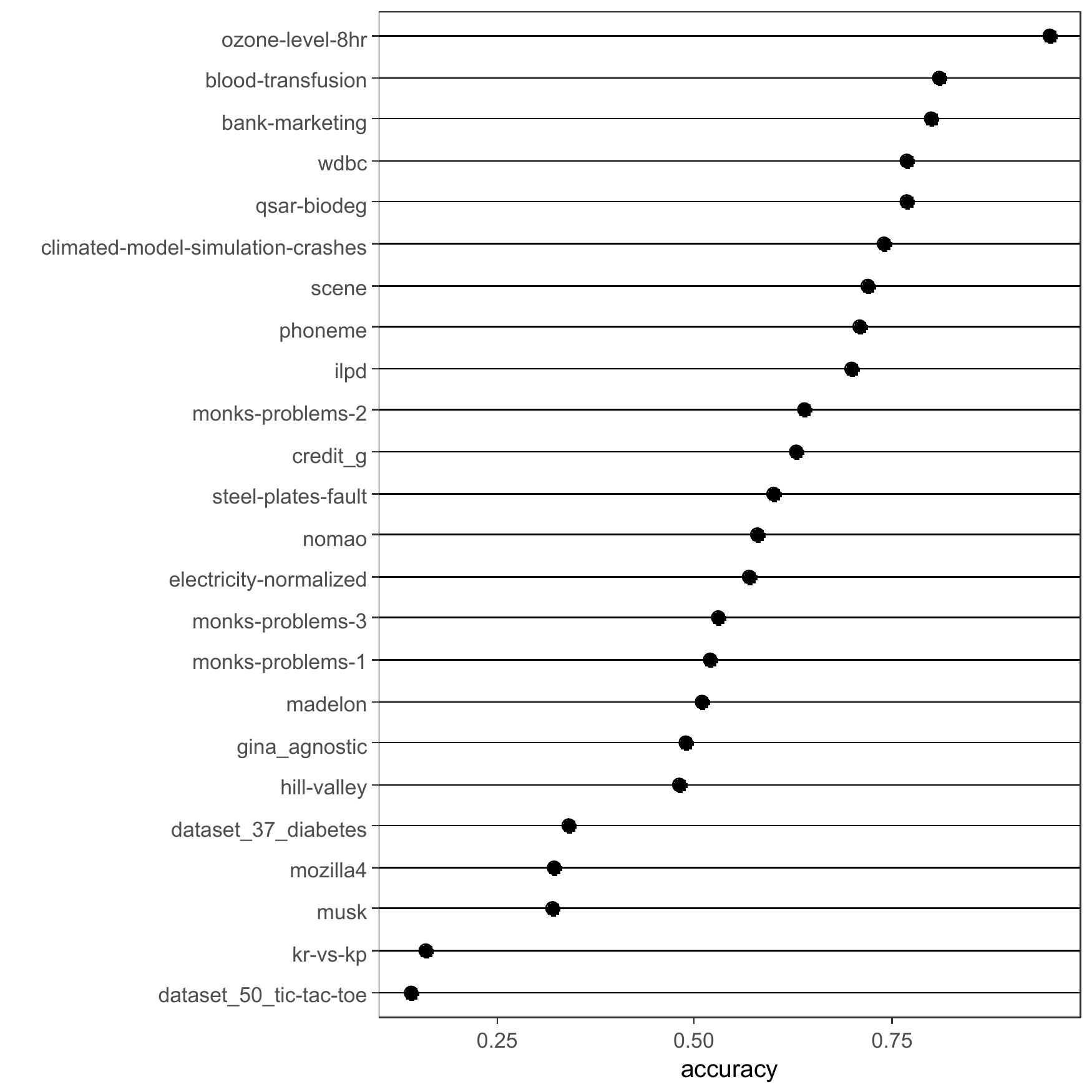}}
			\caption{Grid search results of DBN. }
			\label{dbnGrid}
		\end{subfigure}
		\begin{subfigure}{0.5\textwidth}
			\centering
			\scalebox{0.5}{
				\includegraphics[height=5in]{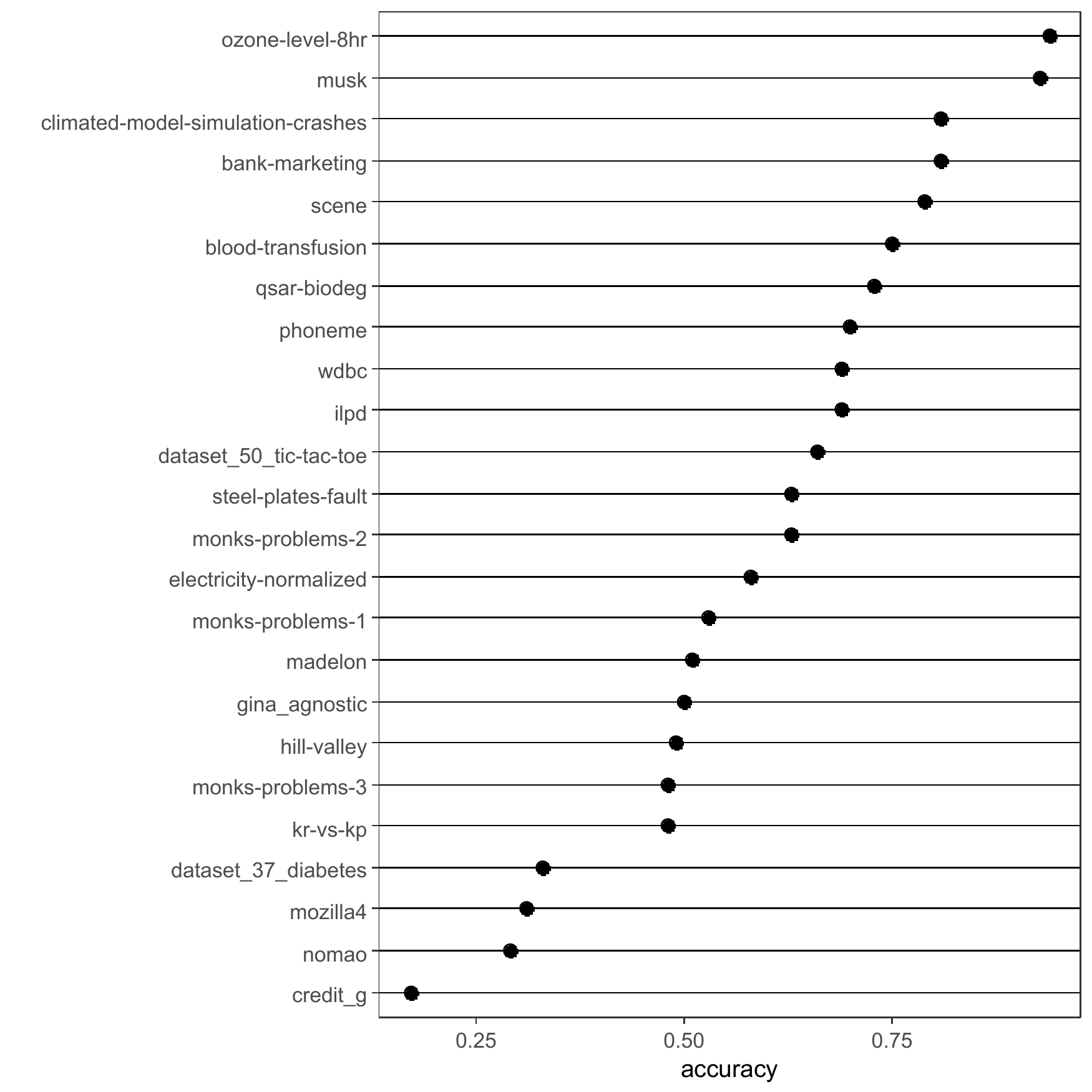}}
			\caption{Random Search results of DBN.}
			\label{dbnRandom}
		\end{subfigure}
		\caption{Accuracy results of DBN.}
		\label{detaildbn}
	\end{figure}

\begin{figure}[H]
		\begin{subfigure}{0.5\textwidth}
			\centering
			\scalebox{0.5}{
				\includegraphics[height=5in]{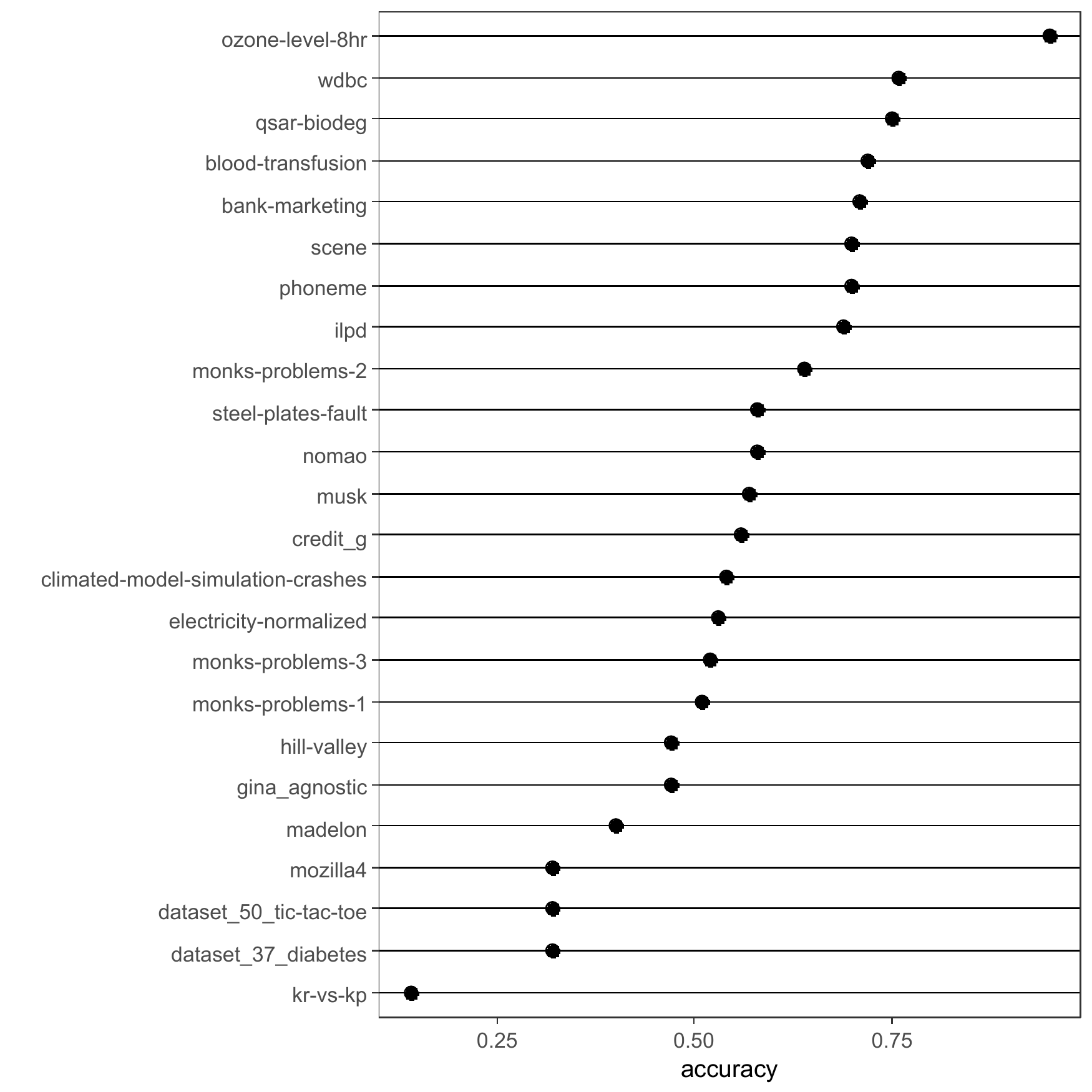}}
			\caption{Grid search results of FFNN. }
			\label{ffnnGrid}
		\end{subfigure}
		\begin{subfigure}{0.5\textwidth}
			\centering
			\scalebox{0.5}{
				\includegraphics[height=5in]{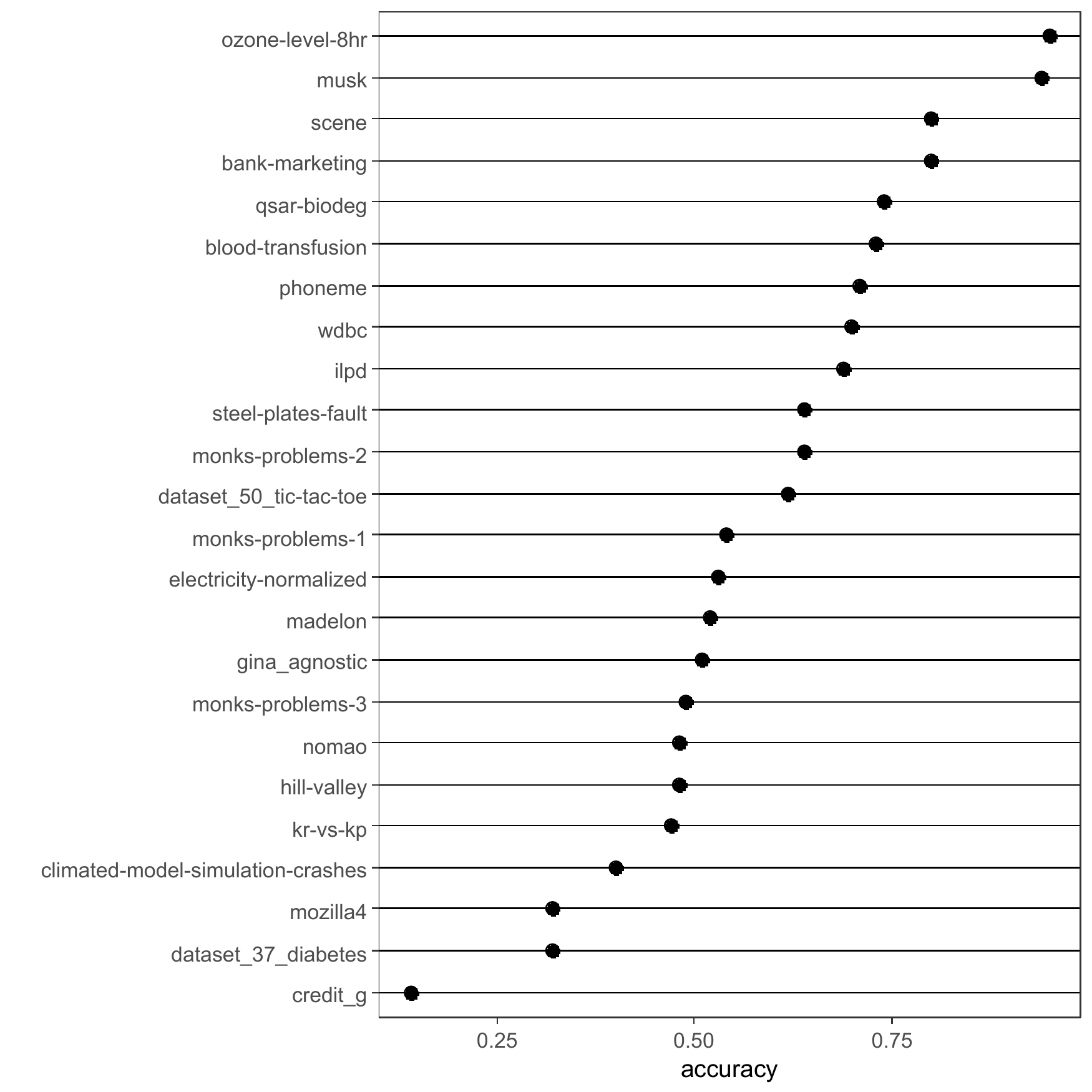}}
			\caption{Random search results of FFNN.}
			\label{ffnnRandom}
		\end{subfigure}
		\caption{Accuracy results of FFNN.}
	\end{figure}

\begin{figure}[H]
		\begin{subfigure}{0.5\textwidth}
			\centering
			\scalebox{0.5}{
				\includegraphics[height=5in]{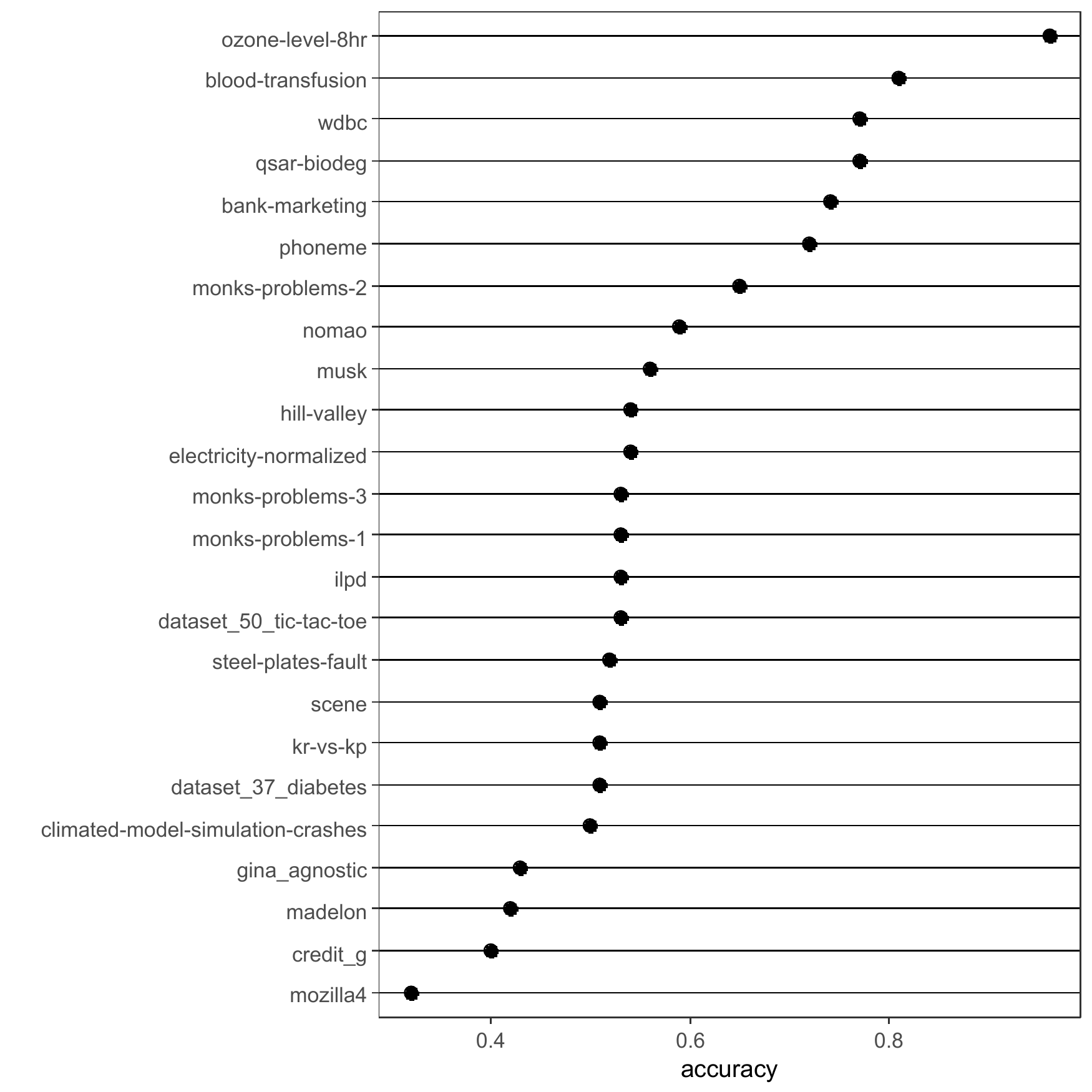}}
			\caption{Grid search results of RNN. }
			\label{rnnGrid}
		\end{subfigure}
		\begin{subfigure}{0.5\textwidth}
			\centering
			\scalebox{0.5}{
				\includegraphics[height=5in]{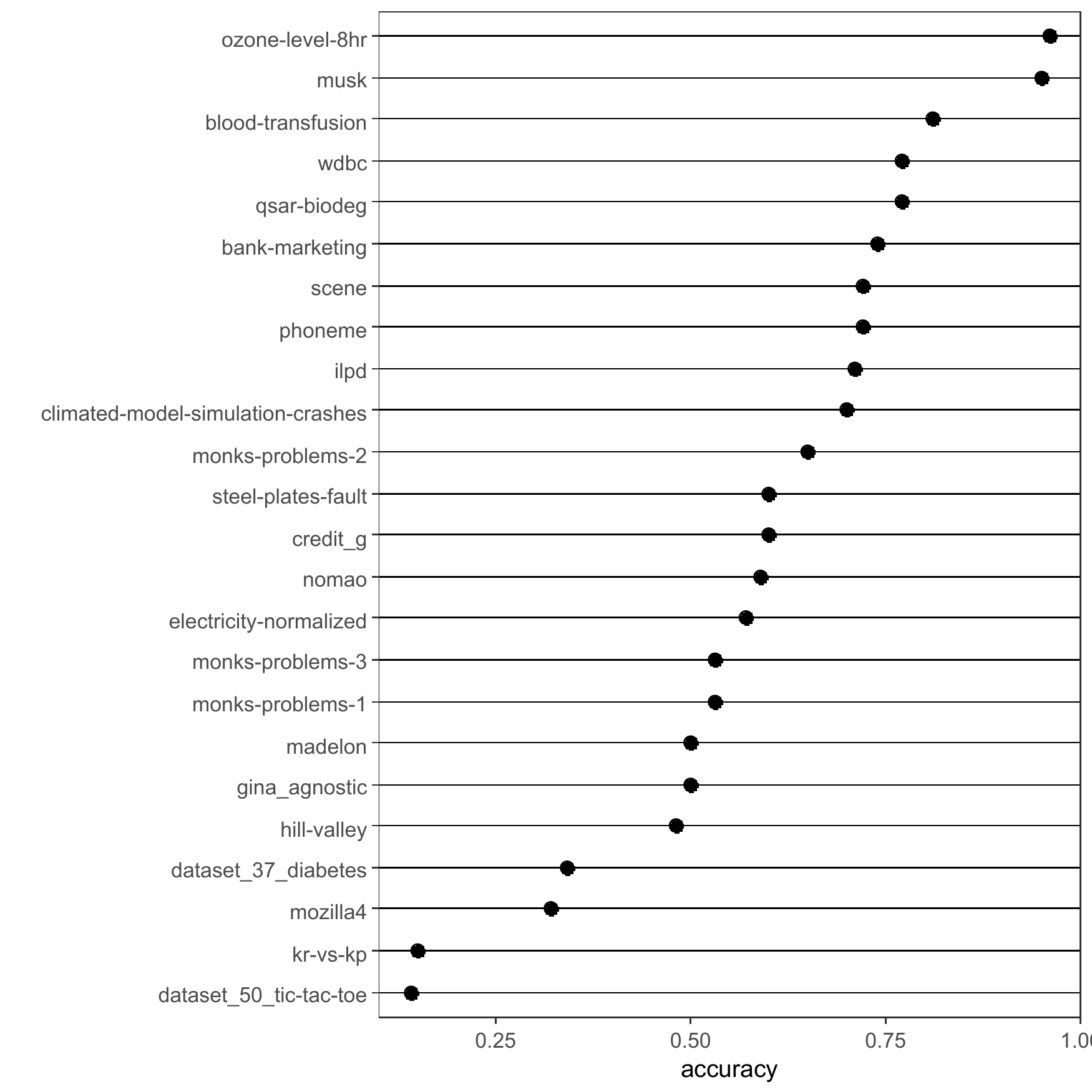}}
			\caption{Random search results of RNN.}
			\label{rnnRandom}
		\end{subfigure}
		\caption{Accuracy results of RNN.}
	\end{figure}

\begin{figure}[H]
		\begin{subfigure}{0.5\textwidth}
			\centering
			\scalebox{0.5}{
				\includegraphics[height=5in]{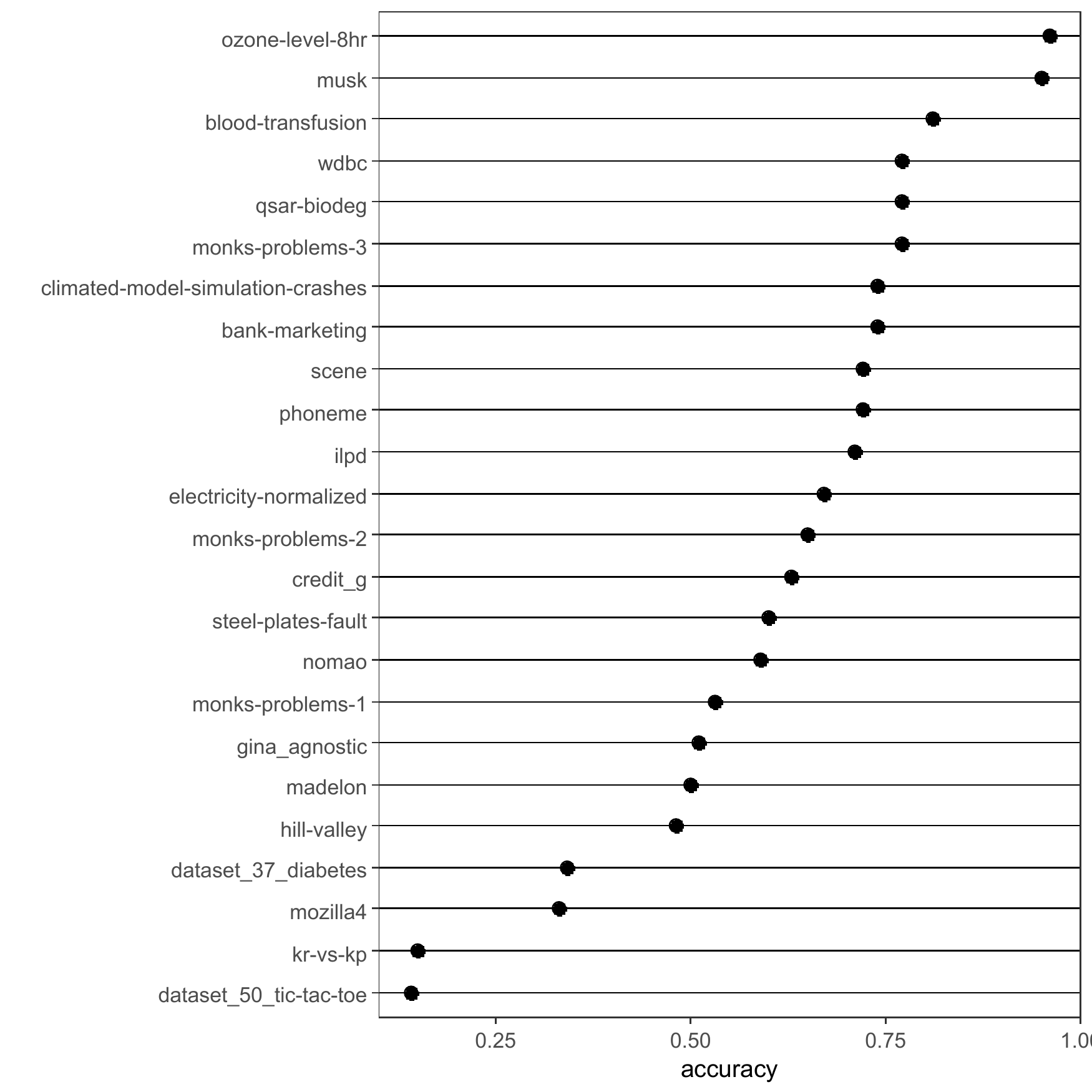}}
			\caption{Grid search results of SAE. }
			\label{saeGrid}
		\end{subfigure}
		\begin{subfigure}{0.5\textwidth}
			\centering
			\scalebox{0.5}{
				\includegraphics[height=5in]{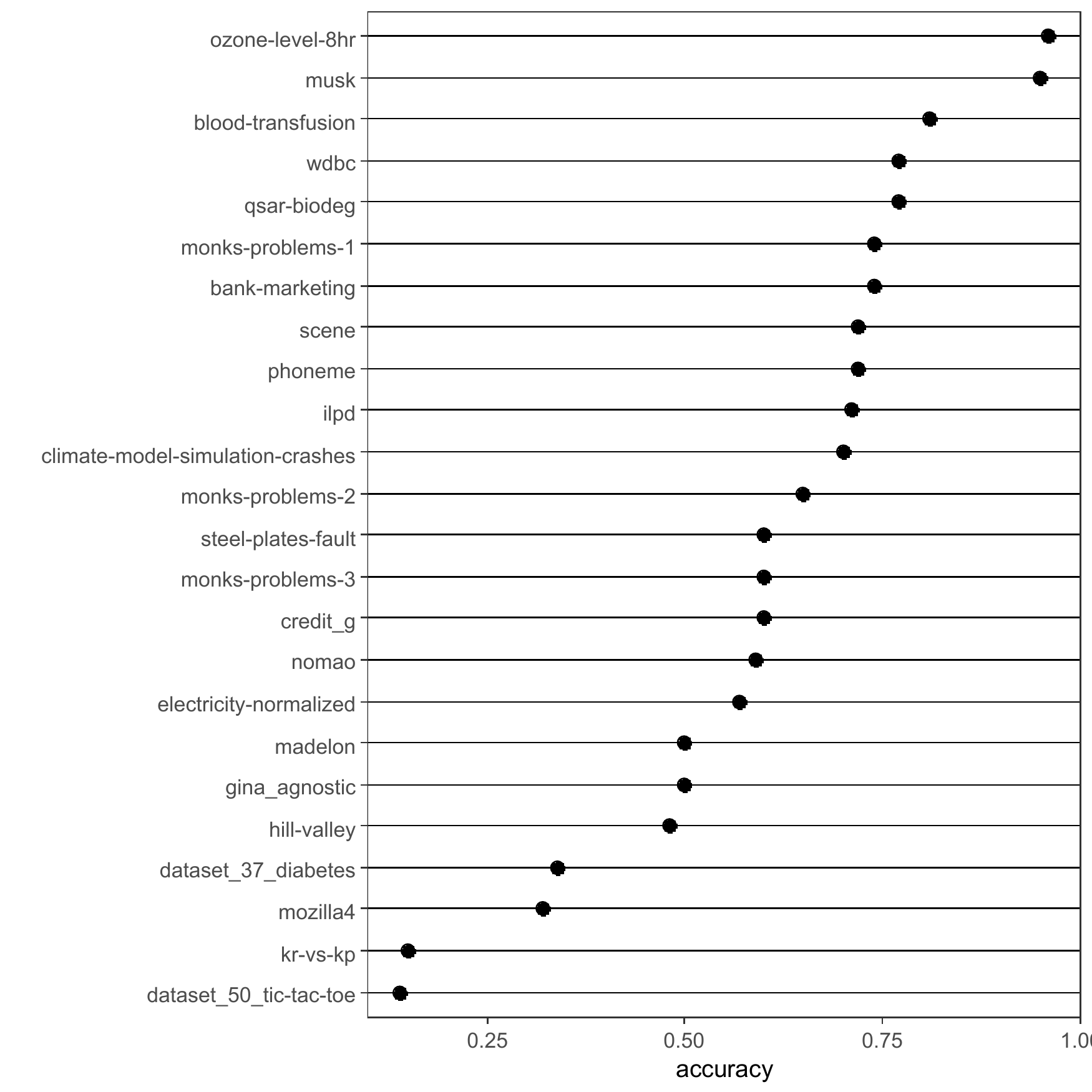}}
			\caption{Random search results of SAE.}
			\label{saeRandom}
		\end{subfigure}
		\caption{Accuracy results of SAE.}
		\label{detailsae}
	\end{figure}

	Detail accuracy results of the data sets are given in Figs. \ref{detaildbn}-\ref{detailsae}. Here the data sets are ranked by accuracy success. It is obviously seen from these figures that ozone-level-8hr has the highest value of accuracy. Low sparsity (0.01229849) may have contributed to the success of ozone-level-8hr. But sparsity is not the sole indicator of the success of hyperparameter search methods. Because, although kr-vs-kp has a low sparsity (0.00189426), it yielded a bad accuracy compared to the other data sets. ozone-level-8hr has a uniform distribution in terms of class labels which is a sign to produce high accuracy. On the other hand, kr-vs-kp has not a uniform distribution in class labels. Further, it completely consists of factor values that were converted to numeric. An interesting result is that the bank-marketing data set has yielded reasonable performance regardless of the type of hyperparameter search method. This data set needs factor to numeric conversion before training. Further, although blood-transfusion has not a high number of features (4), it yielded promising results for all DL algorithms. We can conclude from these results that taking training data from a uniform distribution of class labels is crucial for obtaining high performance for classification. Moreover, the number of features has not a remarkable effect on the success of accuracy. 
	
	\begin{table}
		\setlength{\tabcolsep}{3pt}
		\centering
		\caption{Comparison of the time cost of DL algorithms. Each cell refers to the mean training time of 24 data sets.} 
		\scalebox{1}{
			\begin{tabular}{l rr} 
				\hline 
				
				\textbf{Method} &   \textbf{Time (second)}   \\ [1.5ex]
				\hline 
				DBN & 177\\ 
				DBN-grid search & 789\\ 
				DBN-random search & 215\\ 
				FFNN & 122\\ 
				FFNN-grid search & 850\\ 
				FFNN-random search & 171\\ 
				RNN & 10803\\ 
				RNN-grid search & 57967\\ 
				RNN-random search & 23441\\ 
				SAE & 200\\ 
				SAE-grid search & 1401\\ 
				SAE-random search & 310\\ 
				\hline 
			\end{tabular}
		}
		\label{time}
	\end{table}
	Mean training times of 24 data sets for DL algorithms are given in Table \ref{time}. Inherently, a hyperparameter search method increases the time passed for training a DL algorithm. RNN takes more time (57697sn) than the other algorithms. Elapsed times of DBN are the lowest values of Table \ref{time}. However, if searching hyperparameters is not a must for a DL experiment, FFNN completes the training in minimum time (122sn) compared to the other DL algorithms. 
	\section{Conclusion}
	This paper presented a comprehensive evaluation of some hyperparameters of DL algorithms. The results showed that each hyperparameter should be set in compliance with the design of its associated DL algorithm. High sparsity has a negative effect on the accuracy of DL algorithms. Although DL algorithms have several hyperparameters, the learning rate is the key regardless of the type of DL algorithms. The optimal value of the learning rate mostly depends on the distribution of class labels for classification. 
	
	The experiment performed on 24 classification data sets indicate that normalization has a favorable effect on the increase in accuracy. Further, converting factor values to numeric should be done by considering whether there is a relational pattern among the features having factor values. Hyperparameters of DL algorithms were tuned via a validation set for each data set, thereby changing the optimal hyperparameter set of DL algorithms according to the structure of the data set. We can conclude from the results that disregarding normalization precisely creates a negative impact on the performance of training. Further, evaluating sparsity rates of a data set is strongly related to the class distribution of the data sets that some data sets having zero sparsity produced bad accuracy. On the other hand, having low sparsity can not be considered an obstacle to increase the accuracy where a uniform class distribution is available in the experimental data sets.
	
	There are some ways to extend this work as follows:
	
	1) Derivative-free search methods were not involved in the study. A comparison of derivative-free and blackbox optimization methods may deepen our knowledge about hyperparameter optimization,
	
	2) Using highly-balanced data sets could help improve classification results,
	
	3) Employing an image recognition data set gives a possibility to run an experiment through CNN that may provide new insight into hyperparameter optimization,
	
	4) DL algorithms can be evaluated in a new experimental environment for parallel computation.
	\section*{Acknowledgements}
	We thank TUBITAK ULAKBIM, High Performance and Grid
	Computing Center (TRUBA Resources) for the numerical calculations reported in this
	work.
	
	

	\bigskip
	\rightline{\emph{Received:  {\tiny \raisebox{2pt}{$\bullet$\!}} Revised: }} 
	

\begin{thebibliography}{99}   
		
		{\small
			
			
			
			
			\bibitem{alvarez2016learning} J. M. \href{https://rsu.data61.csiro.au/people/jalvarez/}{Alvarez} and M. Salzmann, Learning the number of neurons in deep networks, \textit{In Advances in Neural Information Processing Systems}, 2016, pp. 2270-2278.
			
			\bibitem{allen1969hugo} G. E. \href{https://education.byu.edu/directory/view/kawika-allen}{Allen}, Hugo de Vries and the reception of the “mutation theory”, \textit{Journal of the History of Biology}, \textbf{2}, 1 (1969) 55-87.
			
			\bibitem{ba2013adaptive} J. \href{http://jimmylba.github.io/}{Ba} and B. Frey, Adaptive dropout for training deep neural networks, \textit{In Advances in neural information processing systems}, 2013, pp. 3084-3092.
			
			\bibitem{baldi2015enhanced} P. \href{http://www.igb.uci.edu/~pfbaldi/}{Baldi}, P. Sadowski, and D. Whiteson, Enhanced Higgs boson to $\tau$+ and $\tau$- search with deep learning, \textit{Phys. Rev. Lett.}, \textbf{114}, 11 (2015) 111801.
			
			\bibitem{bekker2016training} A. J. \href{https://alanbekker.wordpress.com/}{Bekker} and J. Goldberger, Training deep neural-networks based on unreliable labels, \textit{In IEEE International Conference on Acoustics, Speech and Signal Processing (ICASSP)}, 2016, pp. 2682-2686.
			
			\bibitem{bergstra2012random} J. \href{https://www.creativedestructionlab.com/people/james-bergstra/}{Bergstra} and Y. Bengio, Random search for hyper-parameter optimization, \textit{J. Mach. Learn. Res.}, (2012) 281-305.
			
			\bibitem{bjorck2018understanding} N. \href{https://nilsjohanbjorck.github.io/}{Bjorck}, C. P. Gomes, B. Selman, and K. Q. Weinberger, Understanding batch normalization, \textit{In Advances in Neural Information Processing Systems}, 2018, pp. 7694-7705.
			
			\bibitem{pham2014dropout} V. Pham, T. \href{http://www.tbluche.com/}{Bluche}, C. Kermorvant, and J. Louradour, Dropout improves recurrent neural networks for handwriting recognition, \textit{In 14th International Conference on Frontiers in Handwriting Recognition}, 2014, pp. 285-290.
			
			\bibitem{chan2015pcanet}  T. H. \href{http://mx.nthu.edu.tw/~tsunghan/}{Chan}, K. Jia, S. Gao, J. Lu, Z. Zeng, and Y. Ma, PCANet: A simple deep learning baseline for image classification?, \textit{IEEE trans. on image processing}, \textbf{24}, 12 (2015) 5017-5032.
			
			\bibitem{chen2014deep} Y. \href{http://homepage.hit.edu.cn/chenyushi?lang=en}{Chen}, Z. Lin, X. Zhao, G. Wang, and Y. Gu, Deep learning-based classification of hyperspectral data, \textit{IEEE Journal of Selected topics in applied earth observations and remote sensing}, \textbf{7}, 6 (2014) 2094-2107.
			
			\bibitem{choi2018effects} K. \href{https://qcon.ai/qconai2019/speakers/keunwoo-choi}{Choi}, G. Fazekas, K. Cho, and M. Sandler, The effects of noisy labels on deep convolutional neural networks for music tagging, \textit{IEEE Transactions on Emerging Topics in Computational Intelligence}, \textbf{2}, 2 (2018) 139-149.
			
			\bibitem{bhanja2018impact} S. Bhanja and A. \href{https://abhishekdas.com/}{Das}, Impact of data normalization on deep neural network for time series forecasting, \textit{arXiv preprint arXiv:1812.05519}, (2018).
			
			\bibitem{diaz2017effective} G. I. \href{https://www.cs.ox.ac.uk/people/gonzalo.diaz/}{Diaz}, A. Fokoue-Nkoutche, G. Nannicini, and H. Samulowitz, An effective algorithm for hyperparameter optimization of neural networks, \textit{IBM Journal of Research and Development}, \textbf{61}, 4/5 (2017) 9-1.
			
			\bibitem{dobzhansky1950genetic} T. \href{https://tr.wikipedia.org/wiki/Theodosius\_Dobzhansky}{Dobzhansky}, The genetic basis of evolution, \textit{Scientific American}, \textbf{182}, 1 (1950) 32-41.
			
			\bibitem{domhan2015speeding} T. \href{https://www.amazon.science/author/tobias-domhan}{Domhan}, J. T. Springenberg, and F. Hutter, Speeding up automatic hyperparameter optimization of deep neural networks by extrapolation of learning curves, \textit{In Twenty-Fourth International Joint Conference on Artificial Intelligence}, 2015.
			
			\bibitem{duchi2011adaptive} J. \href{https://web.stanford.edu/~jduchi/}{Duchi}, E. Hazan, and Y. Singer, Adaptive subgradient methods for online learning and stochastic optimization. \textit{J. Mach. Learn. Res.}, (2011) 2121-2159.
			
			\bibitem{ensor1997stochastic} K. B. \href{https://statistics.rice.edu/people/katherine-ensor}{Ensor} and P. W. Glynn, Stochastic optimization via grid search. Lectures in Applied Mathematics-American Mathematical Society, \textbf{33}, (1997) 89-100.
			
			\bibitem{fawaz2019deep} H. I \href{https://hfawaz.github.io/}{Fawaz}, G. Forestier, J. Weber, L. Idoumghar, and P. A. Muller, Deep learning for time series classification: a review, \textit{Data Min. Knowl. Discov.}, \textbf{33}, 4 (2019) 917-963.
			
			\bibitem{glaudell1965nelder} R. Glaudell, R. T. \href{https://directorio.uca.es/cau/directorio.do?persona=14365}{Garcia}, and J. B. Garcia, Nelder-mead simplex method, \textit{Computer Journal}, \textbf{7}, 4 (1965) 308-313.
			
			\bibitem{tomar2014towards} A. Tomar, F. \href{https://fredericgodin.com/frederic-godin-machine-learning/}{Godin}, B. Vandersmissen, W. De Neve, and R. Van de Walle, Towards Twitter hashtag recommendation using distributed word representations and a deep feed forward neural network, \textit{In International Conference on Advances in Computing, Communications and Informatics (ICACCI)}, 2014, pp. 362-368.
			
			\bibitem{graves2013speech} A. \href{https://www.cs.toronto.edu/~graves/}{Graves}, A. R. Mohamed, and G. Hinton, Speech recognition with deep recurrent neural networks, \textit{In IEEE international conference on acoustics, speech and signal processing}, 2013, pp. 6645-6649.
			
			\bibitem{guo2016deep} Y. \href{https://scholar.google.com/citations?user=jksxCf8AAAAJ\&hl=zh-CN}{Guo}, Y. Liu, A. Oerlemans, S. Lao, S. Wu, and M. S. Lew, Deep learning for visual understanding: A review, \textit{Neurocomputing}, \textbf{187}, (2016) 27-48. 
			
			\bibitem{held2016learning} D. \href{https://www.ri.cmu.edu/ri-faculty/david-held/}{Held}, S. Thrun, and S. Savarese, Learning to track at 100 fps with deep regression networks, \textit{In European Conference on Computer Vision}, 2016, pp. 749-765.
			
			\bibitem{hinton2006fast} G. E. \href{https://vectorinstitute.ai/team/geoffrey-hinton/}{Hinton}, S. Osindero, and Y. W. Teh, A fast learning algorithm for deep belief nets, \textit{Neural Comput.}, \textbf{18}, 7 (2006) 1527-1554.
			
			\bibitem{hutter2015beyond} F. \href{http://aad.informatik.uni-freiburg.de/people/hutter/}{Hutter}, J. Lücke, and L. Schmidt-Thieme, Beyond manual tuning of hyperparameters, \textit{KI-Künstliche Intelligenz} \textbf{29}, 4 (2015) 329-337.
			
			\bibitem{ilievski2017efficient} I. \href{https://ilija139.github.io}{Ilievski}, T. Akhtar, J. Feng, and C. A. Shoemaker, Efficient hyperparameter optimization for deep learning algorithms using deterministic rbf surrogates, \textit{In Thirty-First AAAI Conference on Artificial Intelligence}, 2017.
			
			\bibitem{keskar2015nonmonotone} N. S. \href{https://keskarnitish.github.io/}{Keskar} and G. Saon, A nonmonotone learning rate strategy for SGD training of deep neural networks, \textit{In IEEE International Conference on Acoustics, Speech and Signal Processing (ICASSP)}, 2015, pp. 4974-4978.
			
			\bibitem{kingma2015variational} D. P. \href{http://dpkingma.com/}{Kingma}, T. Salimans, and M. Welling, Variational dropout and the local reparameterization trick, \textit{In Advances in neural information processing systems}, 2015, pp. 2575-2583.
			
			\bibitem{ko2017controlled} B. \href{https://github.com/kobiso}{Ko}, H. G. Kim, K. J. Oh, and H. J. Choi, Controlled dropout: A different approach to using dropout on deep neural network, \textit{In IEEE International Conference on Big Data and Smart Computing (BigComp)}, 2017, pp. 358-362.
			
			\bibitem{kuri2017closed} A. Kuri-Morales, Closed determination of the number of neurons in the hidden layer of a multi-layered perceptron network, Soft Computing, \textbf{21}, 3 (2017) 597-609.
			
			\bibitem{kussul2017deep} N. \href{http://www.ikd.kiev.ua}{Kussul}, M. Lavreniuk, S. Skakun, and A. Shelestov, Deep learning classification of land cover and crop types using remote sensing data, \textit{IEEE Geoscience and Remote Sensing Letters}, \textbf{14}, 5 (2017) 778-782.
			
			\bibitem{chuck2017statistical} C. Chuck, M. \href{https://people.eecs.berkeley.edu/~laskeymd/}{Laskey}, S. Krishnan, R. Joshi, R. Fox, and K. Goldberg, Statistical data cleaning for deep learning of automation tasks from demonstrations, \textit{In 13th IEEE Conference on Automation Science and Engineering (CASE)}, 2017, pp. 1142-1149.
			
			\bibitem{laurent2016batch} C. \href{https://mila.quebec/en/person/cesar-laurent/}{Laurent}, G. Pereyra, P. Brakel, Y. Zhang, and Y. Bengio, Batch normalized recurrent neural networks, \textit{In IEEE International Conference on Acoustics, Speech and Signal Processing (ICASSP)}, 2016, pp. 2657-2661.
			
			\bibitem{lecun2015deep} Y. \href{http://yann.lecun.com/}{LeCun}, Y. Bengio, Y., and G. Hinton, Deep learning, \textit{nature}, \textbf{521}, 7553 (2015) 436-444.
			
			\bibitem{li2019object} P. \href{https://ieeexplore.ieee.org/author/37276868900}{Li}, X. He, X. Cheng, X. Gao, R. Li, M. Qiao, and Z. Li, Object Extraction From Very High-Resolution Images Using a Convolutional Neural Network Based on a Noisy Large-Scale Dataset, \textit{IEEE Access}, \textbf{7}, (2019), 122784-122795.
			
			\bibitem{li2014efficient} M. \href{https://www.cs.cmu.edu/~muli/}{Li}, T. Zhang, Y. Chen, and A. J. Smola, Efficient mini-batch training for stochastic optimization, \textit{In Proceedings of the 20th ACM SIGKDD international conference on Knowledge discovery and data mining}, 2014, pp. 661-670.
			
			\bibitem{li2018adaptive} Y. \href{https://lyttonhao.github.io/}{Li}, N. Wang, J. Shi, X. Hou, and J. Liu, Adaptive batch normalization for practical domain adaptation, \textit{Pattern Recognition}, \textbf{80}, (2018) 109-117.
			
			\bibitem{li2017hyperband} L. \href{https://liamcli.com/}{Li}, K. Jamieson, G. DeSalvo, A. Rostamizadeh, A. Talwalkar, Hyperband: A novel bandit-based approach to hyperparameter optimization, \textit{The Journal of Machine Learning Research}, \textbf{18}, 1 (2017) 6765-6816.
			
			\bibitem{liu2018deep} M. \href{https://www.linkedin.com/in/mingfeiliu}{Liu}, W. Wu, Z. Gu, Z. Yu, F. Qi, and Y. Li, Deep learning based on Batch Normalization for P300 signal detection, \textit{Neurocomputing}, \textbf{275}, (2018) 288-297.
			
			\bibitem{lopez2019shallow} M. \href{https://www.linkedin.com/in/mlmartin/}{Lopez-Martin}, B. Carro, A. Sanchez-Esguevillas, and J. Lloret, Shallow neural network with kernel approximation for prediction problems in highly demanding data networks, \textit{Expert Systems with Applications}, \textbf{124}, (2019) 196-208.
			
			\bibitem{loshchilov2016cma} I. \href{http://loshchilov.com/index.html}{Loshchilov} and F. Hutter, CMA-ES for hyperparameter optimization of deep neural networks, \textit{arXiv preprint arXiv:1604.07269}, (2016).
			
			\bibitem{massouh2017learning} N. \href{https://phd.uniroma1.it/dottorati/cartellaDocumentiWeb/226902c5-22b0-4d98-8578-67d243e43750.pdf}{Massouh}, F. Babiloni, T. Tommasi, J. Young, N. Hawes, and B. Caputo, Learning deep visual object models from noisy web data: How to make it work, \textit{In IEEE/RSJ International Conference on Intelligent Robots and Systems (IROS)}, pp. 5564-5571.
			
			\bibitem{taylor2018improving} L. Taylor and G. \href{https://www.cs.uct.ac.za/images/staff/geoff.nitschke.jpg/view}{Nitschke}, Improving deep learning with generic data augmentation, \textit{In IEEE Symposium Series on Computational Intelligence (SSCI)}, 2018, pp. 1542-1547.
			
			\bibitem{ozaki2019accelerating} Y. Ozaki, S. \href{https://nabenabe0928.github.io/}{Watanabe}, and M. Onishi, Accelerating the Nelder-Mead method with predictive parallel evaluation, \textit{In 6th ICML Workshop on Automated Machine Learning}, 2019.
			
			\bibitem{ozaki2017effective} Y. \href{https://kyushu-u.pure.elsevier.com/en/persons/yukio-ozaki}{Ozaki}, M. Yano, and M. Onishi, Effective hyperparameter optimization using Nelder-Mead method in deep learning, \textit{IPSJ Transactions on Computer Vision and Applications}, \textbf{9}, 1 (2017) 20.
			
			\bibitem{pal2016preprocessing} K. K. \href{https://kuntalkumarpal.github.io/files/CV.pdf}{Pal} and K. S. Sudeep, Preprocessing for image classification by convolutional neural networks, \textit{In IEEE International Conference on Recent Trends in Electronics, Information \& Communication Technology (RTEICT)}, 2016, pp. 1778-1781.
			
			\bibitem{passalis2019deep} N. \href{https://passalis.github.io/}{Passalis}, A. Tefas, J. Kanniainen, M. Gabbouj, and A. Iosifidis, Deep Adaptive Input Normalization for Time Series Forecasting, \textit{IEEE Transactions on Neural Networks and Learning Systems}, (2019).
			
			\bibitem{wang2017effectiveness} L. \href{https://profiles.stanford.edu/luis-perez-martinez}{Perez} and J. Wang, The effectiveness of data augmentation in image classification using deep learning, \textit{arXiv preprint arXiv:1712.04621}, (2017).
			
			\bibitem{qi2014robust} Y. \href{https://mypage.zju.edu.cn/en/yuqi}{Qi}, Y. Wang, X. Zheng, and Z. Wu, Robust feature learning by stacked autoencoder with maximum correntropy criterion, \textit{In IEEE International Conference on Acoustics, Speech and Signal Processing (ICASSP)}, 2014, pp. 6716-6720.
			
			\bibitem{rios2013derivative} L. M. \href{https://www.e2eanalytics.com/luis-miguel-rios}{Rios} and N. V. Sahinidis, Derivative-free optimization: a review of algorithms and comparison of software implementations, \textit{Journal of Global Optimization}, \textbf{56}, 3 (2013) 1247-1293.
			
			\bibitem{salamon2017deep} J. \href{https://research.adobe.com/person/justin-salamon/}{Salamon} and J. P. Bello, Deep convolutional neural networks and data augmentation for environmental sound classification, \textit{IEEE Signal Processing Letters}, \textbf{24}, 3 (2017) 279-283.
			
			\bibitem{santurkar2018does} S. \href{https://people.csail.mit.edu/shibani/}{Santurkar}, D. Tsipras, A. Ilyas, and A. Madry, How does batch normalization help optimization?, \textit{In Advances in Neural Information Processing Systems}, 2018, pp. 2483-2493.
			
			\bibitem{schmidhuber2015deep} J. \href{https://search.usi.ch/en/people/855dfcba3eaf6e94156db5ff991ba300/schmidhuber-juergen}{Schmidhuber}, Deep learning in neural networks: An overview, Neural networks, \textbf{61}, (2015) 85-117.
			
			\bibitem{scornet2017tuning} E. \href{https://erwanscornet.github.io/}{Scornet}, Tuning parameters in random forests, \textit{ESAIM Proc. Surveys}, \textbf{60}, (2017) 144-162.
			
			\bibitem{chandra2016deep} B. Chandra and R. K. \href{https://in.linkedin.com/in/rajesh-kumar-sharma-8637838}{Sharma}, Deep learning with adaptive learning rate using laplacian score, \textit{Expert Systems with Applications}, \textbf{63}, (2016) 1-7.
			
			\bibitem{singh2015layer} B. \href{https://bharatsingh.net/}{Singh}, S. De, Y. Zhang, T. Goldstein, and G. Taylor, Layer-specific adaptive learning rates for deep networks, \textit{In IEEE 14th International Conference on Machine Learning and Applications (ICMLA)}, 2015, pp. 364-368.
			
			\bibitem{smith2017don} S. L. Smith, P. J. Kindermans, C. Ying, and Q. V. Le, Don't decay the learning rate, increase the batch size, \textit{arXiv preprint arXiv:1711.00489}, (2017).
			
			\bibitem{smith2017natural} E. A. \href{https://faculty.washington.edu/easmith/}{Smith} and B. Winterhalder, Natural selection and decision-making: Some fundamental principles, \textit{In Evolutionary ecology and human behavior}, 2017, pp. 25-60, Routledge.
			
			\bibitem{smith2017cyclical} N. L. \href{https://www.linkedin.com/in/drlesliensmith}{Smith}, Cyclical learning rates for training neural networks, \textit{In IEEE Winter Conference on Applications of Computer Vision (WACV)}, 2017, pp. 464-472.
			
			\bibitem{suk2017deep} H. I. \href{https://koreauniv.pure.elsevier.com/en/persons/heung-il-suk}{Suk}, S. W. Lee, and D. Shen, Deep ensemble learning of sparse regression models for brain disease diagnosis, \textit{Medical image analysis}, \textbf{37}, (2017) 101-113.
			
			\bibitem{sukhbaatar2014learning} S. \href{https://cims.nyu.edu/~sainbar/}{Sukhbaatar} and R. Fergus, Learning from noisy labels with deep neural networks, \textit{arXiv preprint arXiv:1406.2080}, \textbf{2}, 3 (2014) 4.
			
			\bibitem{thomas2016discovery} L. \href{https://in.linkedin.com/in/likewin-thomas-1669881b}{Thomas}, M. Kumar, and B. Annappa, Discovery of optimal neurons and hidden layers in feed-forward Neural Network, \textit{In IEEE International Conference on Emerging Technologies and Innovative Business Practices for the Transformation of Societies (EmergiTech)}, 2016, pp. 286-291.
			
			\bibitem{tran2017bayesian} T. \href{https://www.adelaide.edu.au/directory/toan.m.tran}{Tran}, T. Pham, G. Carneiro, L. Palmer, and I. Reid, A bayesian data augmentation approach for learning deep models, \textit{In Advances in neural information processing systems}, 2017, pp. 2797-2806.
			
			\bibitem{OpenML2013} J. Vanschoren, J. N. Van Rijn, B. Bischl, and L. Torgo, OpenML: networked science in machine learning, \textit{ACM SIGKDD Explorations Newsletter}, \textbf{15}, 2 (2014), 49-60.
			
			\bibitem{vo2017harnessing} P. D. \href{https://vodp.github.io/about/curiculum-vitale.pdf}{Vo}, A. Ginsca, H. Le Borgne, and A. Popescu, Harnessing noisy web images for deep representation., \textit{Computer Vision and Image Understanding}, \textbf{164}, (2017) 68-81.
			
			\bibitem{voulodimos2018deep} A. \href{http://users.teiath.gr/avoulod/}{Voulodimos}, N. \href{http://portal.survey.ntua.gr/main/divisions/topo/doulamis/index.htm}{Doulamis}, A. Doulamis, and E. Protopapadakis, Deep learning for computer vision: A brief review, \textit{Comput. Int. and neuroscience}, (2018).
			
			\bibitem{wang2017deep} Q. \href{https://crabwq.github.io/}{Wang}, J. Wan, and Y. Yuan, Deep metric learning for crowdedness regression, \textit{IEEE Trans. on Circuits and Systems for Video Technology}, \textbf{28}, 10 (2017) 2633-2643.
			
			\bibitem{wang2016novel} G. \href{https://www.uts.edu.au/staff/guoxiu.wang}{Wang}, J. Xu, and B. He, A novel method for tuning configuration parameters of spark based on machine learning. \textit{In 2016 IEEE 18th International Conference on High Performance Computing and Communications}, 2016, pp. 586-593.
			
			\bibitem{probst2019hyperparameters} P. Probst, M. N. \href{https://www.bips-institut.de/en/contact/staff/research-fellow.html?MAid=1110}{Wright}, and A. L. Boulesteix, Hyperparameters and tuning strategies for random forest, \textit{Wiley Interdisciplinary Reviews: Data Mining and Knowledge Discovery}, \textbf{9}, 3 (2019) e1301.
			
			\bibitem{wu2018light} X. \href{https://github.com/AlfredXiangWu}{Wu}, R. He, Z. Sun, and T. Tan, A light cnn for deep face representation with noisy labels, \textit{IEEE Transactions on Information Forensics and Security}, \textbf{13}, 11 2018 2884-2896.
			
			\bibitem{xu2016learning} Q. \href{https://sites.google.com/site/qxuresearchgroup/project-definition}{Xu}, C. Zhang, L. Zhang, and Y. Song, The learning effect of different hidden layers stacked autoencoder, \textit{In 8th International Conference on Intelligent Human-Machine Systems and Cybernetics (IHMSC)}, 2, 2016, pp. 148-151.
			
			\bibitem{yao2018hessian} Z. \href{https://yaozhewei.github.io/}{Yao}, A. Gholami, Q. Lei, K. Keutzer, and M. W. Mahoney, Hessian-based analysis of large batch training and robustness to adversaries, \textit{In Advances in Neural Information Processing Systems}, 2018, pp. 4949-4959.
			
			\bibitem{yaseen2018deep} M. U. \href{https://anjum.web.cern.ch/}{Yaseen}, A. Anjum, O. Rana, and N. Antonopoulos, Deep learning hyper-parameter optimization for video analytics in clouds, \textit{IEEE Transactions on Systems, Man, and Cybernetics: Systems}, \textbf{49}, 1 (2018) 253-264.
			
			\bibitem{yoo2019hyperparameter} Y. Yoo, Hyperparameter optimization of deep neural network using univariate dynamic encoding algorithm for searches, \textit{Knowledge-Based Systems}, \textbf{178}, (2019) 74-83.
			
			\bibitem{young2015optimizing} S. R. \href{https://www.ornl.gov/staff-profile/steven-r-young}{Young}, D. C. Rose, T. P. Karnowski, S. H. Lim, and R. M. Patton, Optimizing deep learning hyper-parameters through an evolutionary algorithm, \textit{In Proceedings of the Workshop on Machine Learning in High-Performance Computing Environments}, 2015, pp. 1-5.
			
			\bibitem{zeiler2012adadelta} M. D. \href{https://www.matthewzeiler.com/}{Zeiler}, Adadelta: an adaptive learning rate method. \textit{arXiv preprint arXiv:1212.5701}, (2012).
			
			\bibitem{zhang2018adaptive} Q. \href{http://cs.stfx.ca/~qzhang/}{Zhang}, L. T. Yang, Z. Chen, P. Li, and F. Bu, An adaptive dropout deep computation model for industrial IoT big data learning with crowdsourcing to cloud computing, \textit{IEEE Transactions on Industrial Informatics}, \textbf{15}, 4 (2018) 2330-2337.
			
			\bibitem{zhang2018generalized} Z. Zhang and H. \href{https://www.ece.cornell.edu/faculty-directory/mert-sabuncu}{Sabuncu}, Generalized cross entropy loss for training deep neural networks with noisy labels, \textit{In Advances in neural information processing systems}, 2018, pp. 8778-8788.
			
			\bibitem{zhao2019research} H. Zhao, F. Liu, H. Zhang, and Z. Liang, Research on a learning rate with energy index in deep learning, \textit{Neural Networks}, \textbf{110}, (2019) 225-231.
			
			\bibitem{zhuo2015adaptive} J. \href{http://ml.cs.tsinghua.edu.cn/~jingwei/}{Zhuo}, J. Zhu, and B. Zhang, Adaptive dropout rates for learning with corrupted features, \textit{In Twenty-Fourth International Joint Conference on Artificial Intelligence}, 2015.
			
		}
	\end{thebibliography}
\end{document}